# Modified LAB Algorithm with Clustering-based Search Space Reduction Method for solving Engineering Design Problems

Ruturaj Reddy, Utkarsh Gupta, Ishaan Kale, Apoorva Shastri, Anand J Kulkarni

Institute of Artificial Intelligence, Dr Vishwanath Karad MIT World Peace University, 124 Paud Road, Kothrud, Pune, MH 411038, India

ruturaj164@gmail.com; 1032201801@mitwpu.edu.in; ishaan.kale@mitwpu.edu.in; apoorva.shastri@mitwpu.edu.in; anand.j.kulkarni@mitwpu.edu.in

**Abstract**

A modified LAB algorithm is introduced in this paper. It builds upon the original LAB algorithm (Reddy et al. 2023), which is a socio-inspired algorithm that models competitive and learning behaviours within a group, establishing hierarchical roles. The proposed algorithm incorporates the roulette wheel approach and a reduction factor introducing inter-group competition and iteratively narrowing down the sample space. The algorithm is validated by solving the benchmark test problems from CEC 2005 and CEC 2017. The solutions are validated using standard statistical tests such as two-sided and pairwise signed rank Wilcoxon test and Friedman rank test. The algorithm exhibited improved and superior robustness as well as search space exploration capabilities. Furthermore, a Clustering-Based Search Space Reduction (C-SSR) method is proposed, making the algorithm capable to solve constrained problems. The C-SSR method enables the algorithm to identify clusters of feasible regions, satisfying the constraints and contributing to achieve the optimal solution. This method demonstrates its effectiveness as a potential alternative to traditional constraint handling techniques. The results obtained using the Modified LAB algorithm are then compared with those achieved by other recent metaheuristic algorithms.

## Introduction

Several nature-inspired optimization algorithms have been developed so far. The notable algorithms include Evolutionary Algorithms (EAs), Genetic Algorithms (GAs), Swarm Optimization (SO) techniques, etc. These methods have proven their superiority in terms of solution quality and computational time over traditional (exact) methods for solving a wide variety of problem classes. When applied to solving a variety of classes of problems, these algorithms require certain modifications as well as some supportive techniques to achieve the global best solution. An Artificial Intelligence (AI) based socio-inspired optimization method, referred to as Leader-Advocate-Believer (LAB) was proposed by Reddy et al. in 2023. The LAB algorithm is inspired by the competitive behavior of individuals in a society. In this algorithm, individuals learn and compete with others within the same group. Once the objective function associated with each individual is evaluated, different roles are assigned based on solution quality. The best solution is referred to as the local leader, the second-best solution as the advocate and the rest as believers. Individuals within a group (i.e., local leader, advocate and believers) compete with one another. Every local leader competes with other local leaders to become the global leader. The advocate and believers in a group compete with one another to become the local leader. This intra and inter-group competition motivates individuals to continuously explore and improve to achieve a global optimal solution. The LAB algorithm was validated on 27 benchmark test problems from CEC 2005, and a statistical comparison using the Wilcoxon-signed rank test was conducted. The results were compared with Particle Swarm Optimization (PSO2011) (Omran MGH, 2011), Covariance Matrix Adaptation (CMAES) (Igel et al., 2006), Artificial Bee Colony (ABC) (Karaboga and Akay, 2009), Self-Adaptive Differential Evolution Algorithm (JDE) (Brest et al., 2006) (SADE) (Qin Yi et al., 2010), Comprehensive Learning Particle Swarm Optimisation (CLPSO) (Liang et al. 2006), Backtracking Search Optimisation Algorithm (BSA) (Civicioglu, 2013) and Ideology Algorithm I(A) (Huan et al., 2017), as well as some recent algorithms such as Whale

Optimization Algorithm (WOA) (Mirjalili and Lewis, 2016), Spotted Hyena Optimizer (SHO) (Dhiman and Kumar, 2017), and African Vulture Optimization Algorithm (AVOA) (Singh et al., 2022). The LAB outperformed most of the algorithms in terms of computational time. Additionally, the algorithm was also validated on 29 benchmark test problems from CEC 2017, and the results were compared with LSHADE-Cn-EpsiN (Awad et al.,2017) and LSHADE (Mohamad et al., 2017). A Friedman test was conducted, where LAB was able to outperform all the algorithms. The LAB algorithm was also capable of solving 23 real-world problems, including AWJM, EDM, parameter tuning of turning titanium alloy, and advanced manufacturing processes problems. The results were then compared with various algorithms such as experimental, regression, FA, variations of CI regression, RSM, FA, BPNN, GA, SA, PSO, and Multi-CI.

It is important to mention here that in the current version of the LAB algorithm (Reddy et al., 2023), individuals other than the local leaders could compete with individuals within the same group only. Even though this helps individuals escape local minima, learning choices are limited since only intra-group learning exists. As a result, the LAB algorithm exhibited weaker exploitation ability. Additionally, it is evident from the results that the LAB algorithm has a higher standard deviation and less robustness. Therefore, there is a need to modify the LAB algorithm to improve the following mechanism and incorporate inter-group competition for every individual.

In this paper, the Modified LAB algorithm is proposed to overcome the challenges mentioned above. An updated follow mechanism is introduced by using the roulette wheel approach, which enables believers to follow advocates from other groups. This helps to develop inter-group competition among individuals and increases their exploration capabilities in the search space. It also increases robustness and reduces the chance of premature convergence. The Modified LAB algorithm also introduces a sampling space reduction factor to iteratively narrow down the search space. The search spaces for leaders and advocates are generated using the sampling space reduction factor, which helps to converge towards promising solutions in higher dimensions. As most real-world problems are constrained in nature, the algorithm must also be capable of solving constrained problems. The modified LAB algorithm for constraint handling is introduced for solving real-world engineering design problems. To achieve this, the Clustering-based Search Space Reduction (C-SSR) method (refer to Section 4.1) for constraints has been devised and incorporated into the Modified LAB. In Modified LAB for solving constrained problems, advocates are allowed to follow leaders from other groups using the roulette wheel approach. At the end of every iteration, all individuals are ranked based on their objective values and then regrouped accordingly. For validation, solutions obtained using the modified LAB are compared with recent metaheuristics.

The paper is organized as follows: Section 2 in detail describes the mathematical formulations and the details about the Modified LAB Algorithm. Section 3 describes the benchmark problems followed by the results and comparison among several algorithms. Section 4 discusses the C-SSR method with an illustrative example followed by constraint handling using Modified LAB. Section 5 describes the engineering design problems, results and comparison with other constraint handling methods. Result analysis and discussion is given in Section 6. The conclusions, notable contributions and a comment on future directions are provided at the end.

## 2  Modified LAB Algorithm

Consider a general optimization problem as follows:

$$Minimize\ f(X) = f(x_1,\ldots,x_i,\ldots,x_N)$$
$$s.t. \quad \psi^l_i \leq x_i \leq \psi_i^u, i = 1,\ldots,N$$

To initiate the algorithm, a population **P** of individuals, $p = 1, \ldots, P$, is generated in the search space $[\psi^l_i, \psi_i^u]$ followed by calculation of their objective function value. These individuals are then randomly assigned to different groups of equal size. **P** represents the total population ($P = n \times G$), where $n$ is number of individuals in each group and $G$ is total number of groups. The algorithm steps are discussed below, and the Modified LAB algorithm flowchart is presented in Fig 1.

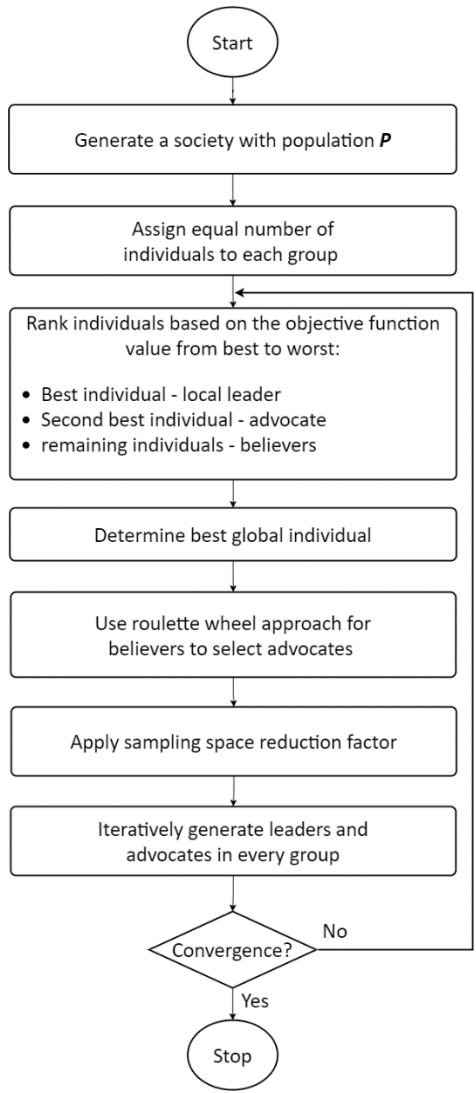

**Fig 1**: Flowchart for Modified LAB

**Step 1: Assigning groups and establishing roles:**

Once assigned to a group, individuals are ranked based on their objective function values. The individual with the best objective function value in each group is referred to as the local leader ($L_g$, ($g = 1, \ldots, G$)), followed by the second-best individual as the advocate ($A_g$, ($g = 1, 2, \ldots, G$)). The remaining individuals ($n - 2$ i.e., excluding the leader and advocate) are referred to as believers ($B_j(g)$, ($j = 1, \ldots, n - 2$)). The local leader with the best objective function value among all the groups is identified as the global leader ($L^*$)

**Step 2: Calculating individual search space using Roulette Wheel Approach**

Once the search space for all individuals is generated, probability-based roulette wheel approach is used, allowing believers to follow advocates not only from their own group but also from other groups.

2.1 The probability of an advocate ($r^{A_g}$) to be selected is calculated as:

$$r^{Ag} = \frac{\frac{1}{f(x)^{Ag}}}{\sum_{g=1}^{G} \frac{1}{f(x)^{Ag}}}$$

2.2 The search space of the believers is influenced by its corresponding leader and the selected advocate, determined through the roulette wheel approach. Additionally, corresponding weights $w_1$ and $w_2$ are randomly chosen, where $w_1 + w_2 = 1$ and $w_1 \geq w_2 \in [0, 1]$. These weights enable partial learning from both the leader and the chosen advocate. This approach ensures that the believers retain their exploration capabilities while benefiting from the knowledge of the leader and advocate. The search space is calculated as follows:

$$\forall x_i^{B_j(g)} \quad x_i^{B_j(g)} = w_1 \times x_i^{L_g} + w_2 \times x_i^{r^{Ag}}, \ \forall B_j(g), i = 1,2,\ldots,N$$

**Step 3: Updating sampling space and generating individual search space for leaders and advocates**

In every iteration, the search space of the local leader and advocate within each group is updated using a sampling space reduction factor of $\theta$

**Step 4: Update the Global and Local Ranking**

Within each group, the individuals (i.e., leader, advocate, and believers) are re-ranked based on the objective function value obtained, and their roles are assigned accordingly. Subsequently, the global leader ($L^*$) is determined based on the objective function values of the local leaders.

**Step 5: Convergence**

If there is no significant improvement in the objective function value of the global leader, or if the maximum iterations have been reached, assume convergence. Else the algorithm proceeds to **Step 2**.

**3 Benchmark Test Problems**

The Modified LAB algorithm is implemented in Python 3 and simulation testing results are generated on Google Colab platform with an Intel(R) Xeon(R) @2.30 GHz Intel Core 2 Duo processor with 12 GB RAM. The algorithm is validated by solving 27 benchmark test functions from CEC 2005 test suite (Table 1), and 29 benchmark test functions from CEC 2017 test suite (Table 2) (Wu et al., 2017). The results obtained for CEC 2005 and CEC 2017 using Modified LAB algorithm are compared with the original LAB algorithm and other contemporary metaheuristic algorithms (refer Table 3 and Table 6). The Wilcoxon signed rank is presented in Table 4 and the pairwise test is presented in Table 5 and Friedman test ranks are presented in Table 7. The following parameters are selected for Modified LAB algorithm based on preliminary trials: individuals $n$ = 5, groups $G$ = 4 and reduction factor $\theta$ = 0.15

**Table 1** CEC 2005 benchmark test suite (Lower Bound (LB) and Upper Bound (UB); S = separable; U = unimodal; N = non-separable; M = multimodal)

| Problem | Name | Type | LB | UB | Dimensions |
|---|---|---|---|---|---|
| F1 | Foxholes | MS | −65.536 | 65.536 | 2 |
| F5 | Ackley | MN | −32 | 32 | 30 |
| F7 | Bohachevsky1 | MS | −100 | 100 | 2 |
| F8 | Bohachevsky2 | MN | −100 | 100 | 2 |
| F9 | Bohachevsky3 | MN | −100 | 100 | 2 |
| F10 | Booth | MS | −10 | 10 | 2 |

| F13 | Dixon-Price | UN | −10 | 10 | 30 |
| F15 | Fletcher | MN | −3.1416 | 3.1416 | 2 |
| F16 | Fletcher | MN | −3.1416 | 3.1416 | 5 |
| F17 | Fletcher | MN | −3.1416 | 3.1416 | 10 |
| F18 | Griewank | MN | −600 | 600 | 30 |
| F19 | Hartman3 | MN | 0 | 1 | 3 |
| F20 | Hartman6 | MN | 0 | 1 | 6 |
| F21 | Kowalik | MN | −5 | 5 | 4 |
| F23 | Langermann5 | MN | 0 | 10 | 5 |
| F24 | Langermann10 | MN | 0 | 10 | 10 |
| F25 | Matyas | UN | −10 | 10 | 2 |
| F32 | Quartic | US | −1.28 | 1.28 | 30 |
| F33 | Rastrigin | MS | −5.12 | 5.12 | 30 |
| F35 | Schaffer | MN | −100 | 100 | 2 |
| F37 | Schwefel_1_2 | UN | −100 | 100 | 30 |
| F38 | Schwefel_2_22 | UN | −10 | 10 | 30 |
| F43 | Six-hump camelback | MN | −5 | 5 | 2 |
| F44 | Sphere2 | US | −100 | 100 | 30 |
| F45 | Step2 | US | −100 | 100 | 30 |
| F47 | Sumsquares | US | −10 | 10 | 30 |
| F50 | Zakharov | UN | −5 | 10 | 10 |

**Table 2** Summary of the CEC 2017 test functions

| Functions | No. | Functions | $F^*_i = F_i(x^*)$ |
|---|---|---|---|
| Unimodal Functions | 1 | Shifted and Rotated Bent Cigar Function | 100 |
| | 2 | Shifted and Rotated Zakharov Function | 200 |
| Simple Multimodal Functions | 3 | Shifted and Rotated Rosenbrock's Function | 300 |
| | 4 | Shifted and Rotated Rastrigin's Function | 400 |
| | 5 | Shifted and Rotated Expanded Scaffer's F6 Function | 500 |
| | 6 | Shifted and Rotated Lunacek Bi_Rastrigin Function | 600 |
| | 7 | Shifted and Rotated Non-Continuous Rastrigin's Function | 700 |
| | 8 | Shifted and Rotated Levy Function | 800 |
| | 9 | Shifted and Rotated Schwefel's Function | 900 |
| Hybrid Functions | 10 | Hybrid Function 1 (N=3) | 1000 |
| | 11 | Hybrid Function 2 (N=3) | 1100 |
| | 12 | Hybrid Function 3 (N=3) | 1200 |
| | 13 | Hybrid Function 4 (N=4) | 1300 |
| | 14 | Hybrid Function 5 (N=4) | 1400 |
| | 15 | Hybrid Function 6 (N=4) | 1500 |
| | 16 | Hybrid Function 6 (N=5) | 1600 |

|  | 17 | Hybrid Function 6 (N=5) | 1700 |
|  | 18 | Hybrid Function 6 (N=5) | 1800 |
|  | 19 | Hybrid Function 6 (N=6) | 1900 |
| Composite Functions | 20 | Composition Function 1 (N=3) | 2000 |
|  | 21 | Composition Function 2 (N=3) | 2100 |
|  | 22 | Composition Function 3 (N=4) | 2200 |
|  | 23 | Composition Function 4 (N=4) | 2300 |
|  | 24 | Composition Function 5 (N=5) | 2400 |
|  | 25 | Composition Function 6 (N=5) | 2500 |
|  | 26 | Composition Function 7 (N=6) | 2600 |
|  | 27 | Composition Function 8 (N=6) | 2700 |
|  | 28 | Composition Function 9 (N=3) | 2800 |
|  | 29 | Composition Function 10 (N=3) | 2900 |
|  |  | Search Range: $[-100, 100]^d$ |  |

**Table 3** Statistical solutions of algorithms for CEC 2005 benchmark problems (Mean = mean solution; Std. Dev. = standard deviation; Best = best solution; Runtime = mean runtime in seconds; NA = not available)

| Function | Statistics | PSO2011 (Clerc, 2011) | CMAES (Igel et al., 2006) | ABC (Karaboga and Akay, 2009) | JDE (Brest et al., 2006) | CLPSO (Liang et al., 2006) | SADE (Qin Yi et al., 2010) | BSA (Civicioglu, 2013) | IA (Huan et al. 2017) | WOA (Seyedali et al., 2016) | SHO (Dhiman et al., 2017) | AVOA (Abdollahzadeh et al., 2021) | LAB (Reddy et al., 2023) | Modified LAB |
|---|---|---|---|---|---|---|---|---|---|---|---|---|---|---|
| F1 | Mean | 1.33E+00 | 1.01E+01 | 9.98E-01 | 1.06E+00 | 1.82E+00 | 9.98E-01 | 9.98E-01 | 9.98E-01 | 2.11E+00 | 9.68E+00 | 1.26E+00 | 9.98E-01 | 0.00E+00 |
|  | Std. Dev. | 9.46E-01 | 8.03E+00 | 1.00E-16 | 3.62E-01 | 1.70E+00 | 0.00E+00 | 0.00E+00 | 3.50E-15 | 2.50E+00 | 3.29E+00 | 5.79E-01 | 0.00E+00 | 0.00E+00 |
|  | Best | 9.98E-01 | 9.98E-01 | 9.98E-01 | 9.98E-01 | 9.98E-01 | 9.98E-01 | 9.98E-01 | 9.98E-01 | NA | NA | 9.98E-01 | 9.98E-01 | 0.00E+00 |
|  | Runtime | 72.53 | 44.79 | 64.98 | 51.10 | 61.65 | 66.63 | 38.13 | 43.54 | NA | NA | NA | 0.05 | 0.32 |
| F5 | Mean | 1.52E+00 | 1.17E+01 | 3.40E-14 | 8.11E-02 | 1.86E-01 | 7.92E-01 | 1.05E-14 | 9.00E-16 | 7.40E+00 | 2.48E+00 | 8.88E-16 | 1.34E+01 | 5.06E-06 |
|  | Std. Dev. | 6.62E-01 | 9.72E+00 | 3.50E-15 | 3.18E-01 | 4.39E-01 | 7.56E-01 | 3.40E-15 | 0.00E+00 | 9.90E+00 | 1.41E+00 | 0.00E+00 | 1.01E+00 | 4.92E-07 |
|  | Best | 8.00E-15 | 8.00E-15 | 2.93E-14 | 4.40E-15 | 8.00E-15 | 4.40E-15 | 8.00E-15 | 9.00E-16 | NA | NA | 8.88E-16 | 1.05E+01 | 4.05E-06 |
|  | Runtime | 63.04 | 3.14 | 23.29 | 11.02 | 45.73 | 40.91 | 14.40 | 49.46 | NA | NA | NA | 1.15 | 1.04 |
| F7 | Mean | 0.00E+00 | 6.22E-02 | 0.00E+00 | 0.00E+00 | 0.00E+00 | 0.00E+00 | 0.00E+00 | 0.00E+00 | NA | NA | NA | 1.63E-01 | 2.40E-11 |
|  | Std. Dev. | 0.00E+00 | 1.35E-01 | 0.00E+00 | 0.00E+00 | 0.00E+00 | 0.00E+00 | 0.00E+00 | 0.00E+00 | NA | NA | NA | 1.70E-01 | 2.31E-11 |
|  | Best | 0.00E+00 | 0.00E+00 | 0.00E+00 | 0.00E+00 | 0.00E+00 | 0.00E+00 | 0.00E+00 | 0.00E+00 | NA | NA | NA | 1.03E-10 | 1.52E-12 |
|  | Runtime | 16.96 | 6.85 | 1.83 | 1.14 | 2.93 | 4.41 | 0.83 | 38.51 | NA | NA | NA | 0.08 | 0.10 |
| F8 | Mean | 0.00E+00 | 7.28E-03 | 0.00E+00 | 0.00E+00 | 0.00E+00 | 0.00E+00 | 0.00E+00 | 0.00E+00 | NA | NA | NA | 1.69E-01 | 2.47E-11 |
|  | Std. Dev. | 0.00E+00 | 3.99E-02 | 0.00E+00 | 0.00E+00 | 0.00E+00 | 0.00E+00 | 0.00E+00 | 0.00E+00 | NA | NA | NA | 2.28E-01 | 2.76E-11 |
|  | Best | 0.00E+00 | 0.00E+00 | 0.00E+00 | 0.00E+00 | 0.00E+00 | 0.00E+00 | 0.00E+00 | 0.00E+00 | NA | NA | NA | 2.22E-08 | 4.77E-15 |
|  | Runtime | 17.04 | 2.17 | 1.80 | 1.14 | 2.89 | 4.42 | 0.82 | 39.02 | NA | NA | NA | 0.08 | 0.10 |
| F9 | Mean | 0.00E+00 | 1.05E-04 | 6.00E-16 | 0.00E+00 | 1.93E-05 | 0.00E+00 | 0.00E+00 | 0.00E+00 | NA | NA | NA | 1.54E-01 | 8.74E-12 |
|  | Std. Dev. | 0.00E+00 | 5.74E-04 | 3.00E-16 | 0.00E+00 | 8.47E-05 | 0.00E+00 | 0.00E+00 | 0.00E+00 | NA | NA | NA | 1.83E-01 | 9.64E-12 |
|  | Best | 0.00E+00 | 0.00E+00 | 1.00E-16 | 0.00E+00 | 0.00E+00 | 0.00E+00 | 0.00E+00 | 0.00E+00 | NA | NA | NA | 0.00E+00 | 2.94E-15 |
|  | Runtime | 17.14 | 2.13 | 21.71 | 1.13 | 33.31 | 4.30 | 0.83 | 40.90 | NA | NA | NA | 0.08 | 0.10 |
| F10 | Mean | 0.00E+00 | 0.00E+00 | 0.00E+00 | 0.00E+00 | 6.01E-04 | 0.00E+00 | 0.00E+00 | 8.35E-01 | NA | NA | NA | 7.02E-06 | 4.77E-01 |

|  |  |  |  |  |  |  |  |  |  |  |  |  |  |
|---|---|---|---|---|---|---|---|---|---|---|---|---|---|
|  | Std. Dev. | 0.00E+00 | 0.00E+00 | 0.00E+00 | 0.00E+00 | 2.99E-03 | 0.00E+00 | 0.00E+00 | 5.00E-16 | NA | NA | NA | 2.10E-05 | 7.04E-01 |
|  | Best | 0.00E+00 | 0.00E+00 | 0.00E+00 | 0.00E+00 | 0.00E+00 | 0.00E+00 | 0.00E+00 | 8.35E-01 | NA | NA | NA | 0.00E+00 | 4.44E-03 |
|  | Runtime | 17.07 | 1.38 | 22.40 | 1.10 | 28.51 | 4.37 | 0.79 | 39.98 | NA | NA | NA | 0.07 | 0.08 |
| **F13** | Mean | 6.67E-01 | 6.67E-01 | 3.80E-15 | 6.67E-01 | 2.33E-03 | 6.67E-01 | 6.44E-01 | 2.53E-01 | NA | NA | NA | 3.75E+04 | 9.83E-01 |
|  | Std. Dev. | 2.20E-15 | 0.00E+00 | 1.20E-15 | 2.00E-16 | 5.18E-03 | 0.00E+00 | 1.22E-01 | 6.51E-10 | NA | NA | NA | 2.14E+04 | 4.68E-03 |
|  | Best | 6.67E-01 | 6.67E-01 | 2.10E-15 | 6.67E-01 | 1.21E-05 | 6.67E-01 | 0.00E+00 | 2.53E-01 | NA | NA | NA | 8.37E+03 | 9.70E-01 |
|  | Runtime | 167.09 | 3.72 | 37.60 | 18.69 | 216.26 | 47.83 | 21.19 | 67.46 | NA | NA | NA | 0.94 | 0.90 |
| **F15** | Mean | 0.00E+00 | 1.03E+03 | 0.00E+00 | 0.00E+00 | 0.00E+00 | 0.00E+00 | 0.00E+00 | 0.00E+00 | NA | NA | NA | 0.00E+00 | 0.00E+00 |
|  | Std. Dev. | 0.00E+00 | 1.30E+03 | 0.00E+00 | 0.00E+00 | 0.00E+00 | 0.00E+00 | 0.00E+00 | 0.00E+00 | NA | NA | NA | 0.00E+00 | 0.00E+00 |
|  | Best | 0.00E+00 | 0.00E+00 | 0.00E+00 | 0.00E+00 | 0.00E+00 | 0.00E+00 | 0.00E+00 | 0.00E+00 | NA | NA | NA | 0.00E+00 | 0.00E+00 |
|  | Runtime | 27.86 | 15.54 | 40.03 | 2.85 | 4.03 | 6.02 | 2.07 | 38.87 | NA | NA | NA | 0.13 | 0.18 |
| **F16** | Mean | 4.87E+01 | 1.68E+03 | 2.19E-02 | 9.44E-01 | 8.18E+01 | 0.00E+00 | 0.00E+00 | 0.00E+00 | NA | NA | NA | 0.00E+00 | 0.00E+00 |
|  | Std. Dev. | 8.89E+01 | 2.45E+03 | 4.18E-02 | 2.88E+00 | 3.80E+02 | 0.00E+00 | 0.00E+00 | 0.00E+00 | NA | NA | NA | 0.00E+00 | 0.00E+00 |
|  | Best | 0.00E+00 | 0.00E+00 | 1.60E-15 | 0.00E+00 | 0.00E+00 | 0.00E+00 | 0.00E+00 | 0.00E+00 | NA | NA | NA | 0.00E+00 | 0.00E+00 |
|  | Runtime | 95.35 | 11.95 | 44.57 | 4.72 | 162.94 | 5.76 | 7.78 | 48.26 | NA | NA | NA | 0.08 | 0.20 |
| **F17** | Mean | 9.19E+02 | 1.23E+04 | 1.11E+01 | 7.14E+02 | 8.53E-01 | 0.00E+00 | 0.00E+00 | 0.00E+00 | NA | NA | NA | 0.00E+00 | 0.00E+00 |
|  | Std. Dev. | 1.65E+03 | 2.24E+04 | 9.88E+00 | 1.71E+03 | 2.92E+00 | 0.00E+00 | 0.00E+00 | 0.00E+00 | NA | NA | NA | 0.00E+00 | 0.00E+00 |
|  | Best | 0.00E+00 | 0.00E+00 | 3.27E-01 | 0.00E+00 | 1.70E-03 | 0.00E+00 | 0.00E+00 | 0.00E+00 | NA | NA | NA | 0.00E+00 | 0.00E+00 |
|  | Runtime | 271.22 | 7.63 | 43.33 | 16.11 | 268.89 | 168.31 | 33.04 | 69.06 | NA | NA | NA | 0.13 | 0.10 |
| **F18** | Mean | 6.89E-03 | 1.15E-03 | 0.00E+00 | 4.82E-03 | 0.00E+00 | 2.26E-02 | 4.93E-04 | 0.00E+00 | 2.89E-04 | 0.00E+00 | 0.00E+00 | 7.86E+01 | 0.00E+00 |
|  | Std. Dev. | 8.06E-03 | 3.64E-03 | 1.00E-16 | 1.33E-02 | 0.00E+00 | 2.84E-02 | 1.88E-03 | 0.00E+00 | 1.59E-03 | 0.00E+00 | 0.00E+00 | 1.76E+01 | 0.00E+00 |
|  | Best | 0.00E+00 | 0.00E+00 | 0.00E+00 | 0.00E+00 | 0.00E+00 | 0.00E+00 | 0.00E+00 | 0.00E+00 | NA | NA | 0.00E+00 | 5.25E+01 | 0.00E+00 |
|  | Runtime | 73.90 | 2.65 | 19.07 | 6.91 | 14.86 | 25.86 | 5.75 | 2.72 | NA | NA | NA | 0.00 | 0.98 |
| **F19** | Mean | -3.86E+00 | -3.72E+00 | -3.86E+00 | -3.86E+00 | -3.86E+00 | -3.86E+00 | -3.86E+00 | -3.86E+00 | -3.86E+00 | -3.75E+00 | -3.86E+00 | -3.83E+00 | -3.71E+00 |
|  | Std. Dev. | 2.70E-15 | 5.41E-01 | 2.40E-15 | 2.70E-15 | 2.70E-15 | 2.70E-15 | 2.70E-15 | 3.40E-03 | 2.71E-03 | 4.39E-01 | 9.16E-10 | 1.92E-02 | 1.26E-01 |
|  | Best | -3.86E+00 | -3.86E+00 | -3.86E+00 | -3.86E+00 | -3.86E+00 | -3.86E+00 | -3.86E+00 | -3.86E+00 | NA | NA | -3.86E+00 | -3.86E+00 | -3.85E+00 |
|  | Runtime | 19.28 | 21.88 | 12.61 | 7.51 | 17.50 | 24.80 | 6.01 | 46.17 | NA | NA | NA | 0.13 | 0.17 |
| **F20** | Mean | -3.32E+00 | -3.29E+00 | -3.32E+00 | -3.30E+00 | -3.32E+00 | -3.31E+00 | -3.32E+00 | -2.57E+00 | -2.98E+00 | -1.44E+00 | -3.31E+00 | -3.32E+00 | -3.85E+00 |
|  | Std. Dev. | 2.17E-02 | 5.11E-02 | 1.40E-15 | 4.84E-02 | 1.30E-15 | 3.02E-02 | 1.30E-15 | 0.00E+00 | 3.77E-01 | 5.47E-01 | 3.02E-02 | 3.45E-02 | 8.48E-02 |
|  | Best | -3.32E+00 | -3.32E+00 | -3.32E+00 | -3.32E+00 | -3.32E+00 | -3.32E+00 | -3.32E+00 | -2.57E+00 | NA | NA | -3.32E+00 | -3.37E+00 | -3.96E+00 |
|  | Runtime | 26.21 | 7.33 | 13.56 | 8.01 | 20.10 | 33.72 | 6.82 | 59.08 | NA | NA | NA | 0.22 | 0.28 |
| **F21** | Mean | 3.07E-04 | 6.48E-03 | 4.41E-04 | 3.69E-04 | 3.10E-04 | 3.07E-04 | 3.07E-04 | 1.70E-03 | 5.72E-04 | 9.01E-04 | 4.65E-04 | 5.13E-02 | 3.91E-02 |
|  | Std. Dev. | 0.00E+00 | 1.49E-02 | 5.68E-05 | 2.32E-04 | 5.98E-06 | 0.00E+00 | 0.00E+00 | 1.31E-06 | 3.24E-04 | 1.06E-04 | 1.48E-04 | 2.37E-02 | 4.84E-03 |
|  | Best | 3.07E-04 | 3.07E-04 | 3.23E-04 | 3.07E-04 | 3.07E-04 | 3.07E-04 | 3.07E-04 | 1.70E-03 | NA | NA | 3.08E-04 | 1.38E-02 | 2.50E-02 |
|  | Runtime | 84.47 | 13.86 | 20.26 | 7.81 | 156.10 | 45.44 | 11.72 | 48.92 | NA | NA | NA | 0.17 | 0.21 |
| **F23** | Mean | -1.39E+00 | -5.24E-01 | -1.50E+00 | -1.34E+00 | -1.48E+00 | -1.50E+00 | -1.48E+00 | -1.50E+00 | NA | NA | NA | -9.41E-01 | -1.50E+00 |
|  | Std. Dev. | 2.26E-01 | 2.59E-01 | 8.44E-07 | 2.68E-01 | 1.28E-01 | 9.00E-16 | 9.77E-02 | 0.00E+00 | NA | NA | NA | 3.08E-01 | 4.25E-14 |
|  | Best | -1.50E+00 | -7.98E-01 | -1.50E+00 | -1.50E+00 | -1.50E+00 | -1.50E+00 | -1.50E+00 | -1.50E+00 | NA | NA | NA | -1.50E+00 | -1.50E+00 |
|  | Runtime | 33.81 | 17.94 | 37.99 | 20.33 | 42.49 | 36.04 | 18.93 | 41.85 | NA | NA | NA | 0.54 | 0.60 |
| **F24** | Mean | -9.17E-01 | -3.11E-01 | -8.41E-01 | -8.83E-01 | -9.43E-01 | -1.28E+00 | -1.31E+00 | -1.50E+00 | NA | NA | NA | -2.33E-01 | -1.50E+00 |
|  | Std. Dev. | 3.92E-01 | 2.08E-01 | 2.00E-01 | 3.88E-01 | 3.18E-01 | 3.60E-01 | 3.16E-01 | 0.00E+00 | NA | NA | NA | 2.38E-01 | 8.02E-14 |
|  | Best | -1.50E+00 | -7.98E-01 | -1.50E+00 | -1.50E+00 | -1.50E+00 | -1.50E+00 | -1.50E+00 | -1.50E+00 | NA | NA | NA | -7.98E-01 | -1.50E+00 |
|  | Runtime | 110.80 | 8.84 | 38.47 | 21.60 | 124.61 | 47.17 | 35.36 | 54.65 | NA | NA | NA | 1.09 | 1.14 |
| **F25** | Mean | 0.00E+00 | 0.00E+00 | 4.00E-16 | 0.00E+00 | 4.18E-06 | 0.00E+00 | 0.00E+00 | 0.00E+00 | NA | NA | NA | 2.00E-16 | 1.50E-15 |
|  | Std. Dev. | 0.00E+00 | 0.00E+00 | 3.00E-16 | 0.00E+00 | 1.62E-05 | 0.00E+00 | 0.00E+00 | 0.00E+00 | NA | NA | NA | 1.00E-15 | 2.63E-15 |
|  | Best | 0.00E+00 | 0.00E+00 | 1.00E-16 | 0.00E+00 | 0.00E+00 | 0.00E+00 | 0.00E+00 | 0.00E+00 | NA | NA | NA | 0.00E+00 | 2.99E-19 |
|  | Runtime | 25.36 | 1.34 | 19.69 | 1.14 | 31.63 | 4.09 | 0.81 | 35.66 | NA | NA | NA | 0.07 | 0.08 |
| **F32** | Mean | 3.55E-04 | 7.02E-02 | 2.50E-02 | 1.30E-03 | 1.96E-03 | 1.67E-03 | 2.00E-03 | 2.25E-04 | 1.43E-03 | 1.06E-04 | 2.52E-04 | 2.58E+00 | 5.76E-27 |
|  | Std. Dev. | 1.41E-04 | 2.89E-02 | 7.72E-03 | 9.95E-04 | 4.34E-03 | 7.33E-04 | 9.70E-04 | 5.27E-04 | 1.15E-03 | 2.43E-05 | 2.05E-04 | 1.38E+00 | 1.84E-27 |
|  | Best | 1.01E-04 | 2.99E-02 | 9.46E-03 | 1.79E-04 | 4.21E-04 | 5.63E-04 | 6.08E-04 | 2.38E-06 | NA | NA | 4.26E-06 | 6.54E-01 | 2.62E-27 |
|  | Runtime | 290.67 | 2.15 | 34.98 | 82.12 | 103.28 | 171.64 | 48.24 | 218.72 | NA | NA | NA | 0.86 | 0.77 |
| **F33** | Mean | 2.56E+01 | 9.60E+01 | 0.00E+00 | 1.13E+00 | 6.30E-01 | 8.62E-01 | 0.00E+00 | 0.00E+00 | 0.00E+00 | 0.00E+00 | 0.00E+00 | 2.56E+02 | 2.48E-10 |

|  |  |  |  |  |  |  |  |  |  |  |  |  |  |
|---|---|---|---|---|---|---|---|---|---|---|---|---|---|
|  | Std. Dev. | 8.29E+00 | 5.67E+01 | 0.00E+00 | 1.07E+00 | 8.05E-01 | 9.32E-01 | 0.00E+00 | 0.00E+00 | 0.00E+00 | 0.00E+00 | 0.00E+00 | 3.11E+01 | 4.90E-11 |
|  | Best | 1.29E+01 | 2.98E+01 | 0.00E+00 | 0.00E+00 | 0.00E+00 | 0.00E+00 | 0.00E+00 | 0.00E+00 | NA | NA | 0.00E+00 | 1.99E+02 | 1.18E-10 |
|  | Runtime | 76.08 | 2.74 | 4.09 | 7.64 | 18.43 | 23.59 | 5.40 | 2.27 | NA | NA | NA | 0.97 | 1.02 |
| **F35** | Mean | 0.00E+00 | 4.65E-01 | 0.00E+00 | 3.89E-03 | 1.94E-03 | 6.48E-04 | 0.00E+00 | 0.00E+00 | NA | NA | NA | 5.80E-03 | 1.19E-15 |
|  | Std. Dev. | 0.00E+00 | 9.34E-02 | 0.00E+00 | 4.84E-03 | 3.95E-03 | 2.47E-03 | 0.00E+00 | 0.00E+00 | NA | NA | NA | 1.29E-02 | 1.47E-15 |
|  | Best | 0.00E+00 | 9.72E-03 | 0.00E+00 | 0.00E+00 | 0.00E+00 | 0.00E+00 | 0.00E+00 | 0.00E+00 | NA | NA | NA | 1.83E-09 | 0.00E+00 |
|  | Runtime | 18.16 | 24.02 | 7.86 | 4.22 | 8.30 | 5.90 | 1.78 | 33.16 | NA | NA | NA | 0.08 | 0.10 |
| **F37** | Mean | 0.00E+00 | 0.00E+00 | 1.46E+01 | 0.00E+00 | 6.47E+00 | 0.00E+00 | 0.00E+00 | 0.00E+00 | 5.39E-07 | 0.00E+00 | 7.83E-145 | 1.44E+04 | 4.23E-13 |
|  | Std. Dev. | 0.00E+00 | 0.00E+00 | 8.71E+00 | 0.00E+00 | 8.22E+00 | 0.00E+00 | 0.00E+00 | 0.00E+00 | 2.93E-06 | 0.00E+00 | 4.29E-144 | 4.22E+03 | 4.27E-13 |
|  | Best | 0.00E+00 | 0.00E+00 | 4.04E+00 | 0.00E+00 | 1.82E-01 | 0.00E+00 | 0.00E+00 | 0.00E+00 | NA | NA | 4.92E-217 | 6.90E+03 | 7.63E-16 |
|  | Runtime | 543.18 | 3.37 | 111.84 | 19.31 | 179.08 | 109.55 | 57.29 | 100.95 | NA | NA | NA | 1.50 | 0.15 |
| **F38** | Mean | 0.00E+00 | 0.00E+00 | 5.00E-16 | 0.00E+00 | 0.00E+00 | 0.00E+00 | 0.00E+00 | 0.00E+00 | 1.06E-21 | 0.00E+00 | 8.72E-104 | 3.77E+00 | 9.80E-06 |
|  | Std. Dev. | 0.00E+00 | 0.00E+00 | 1.00E-16 | 0.00E+00 | 0.00E+00 | 0.00E+00 | 0.00E+00 | 0.00E+00 | 2.39E-21 | 0.00E+00 | 4.71E-103 | 8.67E-01 | 9.17E-07 |
|  | Best | 0.00E+00 | 0.00E+00 | 3.00E-16 | 0.00E+00 | 0.00E+00 | 0.00E+00 | 0.00E+00 | 0.00E+00 | NA | NA | 2.87E-136 | 2.19E+00 | 7.90E-06 |
|  | Runtime | 163.19 | 2.56 | 20.59 | 1.49 | 12.56 | 5.63 | 3.21 | 47.01 | NA | NA | NA | 0.88 | 0.78 |
| **F43** | Mean | -1.03E+00 | -1.00E+00 | -1.03E+00 | -1.03E+00 | -1.03E+00 | -1.03E+00 | -1.03E+00 | -1.03E+00 | -1.03E+00 | -1.06E+01 | -1.03E+00 | -1.03E+00 | -1.01E+00 |
|  | Std. Dev. | 5.00E-16 | 1.49E-01 | 0.00E+00 | 0.00E+00 | 5.00E-16 | 5.00E-16 | 5.00E-16 | 1.49E-03 | 4.20E-07 | 2.86E-11 | 6.78E-16 | 2.25E-03 | 2.42E-02 |
|  | Best | -1.03E+00 | -1.03E+00 | -1.03E+00 | -1.03E+00 | -1.03E+00 | -1.03E+00 | -1.03E+00 | -1.03E+00 | NA | NA | -1.03E+00 | -1.03E+00 | -1.03E+00 |
|  | Runtime | 16.75 | 24.80 | 11.31 | 7.15 | 18.56 | 27.65 | 5.69 | 39.90 | NA | NA | NA | 0.09 | 0.10 |
| **F44** | Mean | 0.00E+00 | 0.00E+00 | 0.00E+00 | 0.00E+00 | 0.00E+00 | 0.00E+00 | 0.00E+00 | 0.00E+00 | 1.41E-30 | 0.00E+00 | 2.01E-199 | 9.19E+03 | 2.51E-11 |
|  | Std. Dev. | 0.00E+00 | 0.00E+00 | 0.00E+00 | 0.00E+00 | 0.00E+00 | 0.00E+00 | 0.00E+00 | 0.00E+00 | 4.91E-30 | 0.00E+00 | 0.00E+00 | 1.60E+03 | 5.85E-11 |
|  | Best | 0.00E+00 | 0.00E+00 | 0.00E+00 | 0.00E+00 | 0.00E+00 | 0.00E+00 | 0.00E+00 | 0.00E+00 | NA | NA | 5.44E-269 | 6.52E+03 | 2.85E-16 |
|  | Runtime | 159.90 | 2.32 | 21.92 | 1.42 | 14.39 | 5.92 | 3.30 | 174.58 | NA | NA | NA | 0.87 | 0.70 |
| **F45** | Mean | 2.30E+00 | 6.67E-02 | 0.00E+00 | 9.00E-01 | 0.00E+00 | 0.00E+00 | 0.00E+00 | 5.39E-05 | 3.12E+00 | 2.46E-01 | 2.43E-06 | 8.20E+03 | 0.00E+00 |
|  | Std. Dev. | 1.86E+00 | 2.54E-01 | 0.00E+00 | 3.02E+00 | 0.00E+00 | 0.00E+00 | 0.00E+00 | 5.40E-10 | 5.32E-01 | 1.78E-01 | 1.89E-06 | 1.96E+03 | 0.00E+00 |
|  | Best | 0.00E+00 | 0.00E+00 | 0.00E+00 | 0.00E+00 | 0.00E+00 | 0.00E+00 | 0.00E+00 | 5.39E-05 | NA | NA | 2.43E-07 | 5.21E+03 | 0.00E+00 |
|   | Runtime  | 57.28 | 1.48 | 1.78 | 2.92 | 3.04 | 4.31 | 0.88 | 2.22 | NA | NA | NA | 0.87 | 0.78 |
| F47 | Mean     | 0.00E+00 | 0.00E+00 | 0.00E+00 | 0.00E+00 | 0.00E+00 | 0.00E+00 | 0.00E+00 | 0.00E+00 | NA | NA | NA | 1.22E+03 | 7.68E-12 |
|   | Std. Dev.| 0.00E+00 | 0.00E+00 | 0.00E+00 | 0.00E+00 | 0.00E+00 | 0.00E+00 | 0.00E+00 | 0.00E+00 | NA | NA | NA | 3.37E+02 | 1.14E-11 |
|   | Best     | 0.00E+00 | 0.00E+00 | 0.00E+00 | 0.00E+00 | 0.00E+00 | 0.00E+00 | 0.00E+00 | 0.00E+00 | NA | NA | NA | 5.59E+02 | 1.86E-14 |
|   | Runtime  | 564.18 | 2.57 | 24.17 | 1.87 | 15.95 | 6.38 | 4.31 | 31.30 | NA | NA | NA | 0.88 | 0.77 |
| F50 | Mean     | 0.00E+00 | 0.00E+00 | 4.02E-08 | 0.00E+00 | 1.60E-10 | 0.00E+00 | 0.00E+00 | 0.00E+00 | NA | NA | NA | 2.33E+04 | 1.69E-12 |
|   | Std. Dev.| 0.00E+00 | 0.00E+00 | 2.20E-07 | 0.00E+00 | 6.27E-10 | 0.00E+00 | 0.00E+00 | 0.00E+00 | NA | NA | NA | 8.02E+04 | 6.39E-13 |
|   | Best     | 0.00E+00 | 0.00E+00 | 2.10E-14 | 0.00E+00 | 0.00E+00 | 0.00E+00 | 0.00E+00 | 0.00E+00 | NA | NA | NA | 1.49E+01 | 6.94E-13 |
|   | Runtime  | 86.37 | 1.87 | 86.45 | 1.41 | 157.84 | 4.93 | 5.70 | 33.57 | NA | NA | NA | 0.33 | 0.30 |

Table 4 Statistical results for CEC 2005 benchmark test problems using two-sided Wilcoxon signed rank test ($\alpha$ = 0.05)

| Test Functions | CLPSO vs Modified LAB | | | | SADE vs Modified LAB | | | | BSA vs Modified LAB | | | | IA vs Modified LAB | | | |
|---|---|---|---|---|---|---|---|---|---|---|---|---|---|---|---|---|
|   | P-value | T+ | T- | Winner | P-value | T+ | T- | Winner | P-value | T+ | T- | Winner | P-value | T+ | T- | Winner |
| F1  | 4.3205E-08 | 0   | 465 | + | 1          | 0   | 0   | = | 1          | 0   | 0   | = | 1          | 0   | 0   | = |
| F5  | 1.7311E-06 | 0   | 465 | + | 1.7311E-06 | 0   | 465 | + | 1.7311E-06 | 465 | 0   | - | 1.7311E-06 | 465 | 0   | - |
| F7  | 1          | 0   | 0   | = | 1          | 0   | 0   | = | 1          | 0   | 0   | = | 1          | 0   | 0   | = |
| F8  | 1          | 0   | 0   | = | 1          | 0   | 0   | = | 1          | 0   | 0   | = | 1          | 0   | 0   | = |
| F9  | 4.3205E-08 | 0   | 465 | + | 1          | 0   | 0   | = | 1          | 0   | 0   | = | 4.3205E-08 | 0   | 465 | + |
| F10 | 1.9209E-06 | 464 | 1   | - | 1.7344E-06 | 465 | 0   | - | 1.7311E-06 | 465 | 0   | - | 2.4118E-04 | 54  | 411 | + |
| F13 | 1.6901E-06 | 465 | 0   | - | 1.6901E-06 | 465 | 0   | - | 1.6901E-06 | 465 | 0   | - | 4.069E-05  | 432 | 33  | - |
| F15 | 1          | 0   | 0   | = | 1          | 0   | 0   | = | 1          | 0   | 0   | = | 1          | 0   | 0   | = |
| F16 | 4.3205E-08 | 0   | 465 | + | 1          | 0   | 0   | = | 1          | 0   | 0   | = | 1          | 0   | 0   | = |
| F17 | 4.3205E-08 | 0   | 465 | + | 1          | 0   | 0   | = | 1          | 0   | 0   | = | 1          | 0   | 0   | = |

| | | | | | | | | | | | | | | | |
|---|---|---|---|---|---|---|---|---|---|---|---|---|---|---|---|
| F18 | 1 | 0 | 0 | = | 4.3205E-08 | 0 | 465 | + | 4.3205E-08 | 0 | 465 | + | 1 | 0 | 0 | = |
| F19 | 1.6044E-06 | 465 | 0 | - | 1.604E-06 | 465 | 0 | - | 1.6044E-06 | 465 | 0 | - | 1.7105E-06 | 465 | 0 | - |
| F20 | 1.7127E-06 | 0 | 465 | + | 1.7127E-06 | 0 | 465 | + | 1.7127E-06 | 0 | 465 | + | 1.7127E-06 | 0 | 465 | + |
| F21 | 1.7213E-06 | 465 | 0 | - | 1.7213E-06 | 465 | 0 | - | 1.7213E-06 | 465 | 0 | - | 1.7224E-06 | 465 | 0 | - |
| F23 | 4.3205E-08 | 0 | 465 | + | 4.3205E-08 | 0 | 465 | + | 4.3205E-08 | 0 | 465 | + | 1 | 0 | 1 | + |
| F24 | 4.3205E-08 | 0 | 465 | + | 4.3205E-08 | 0 | 465 | + | 4.3205E-08 | 0 | 465 | + | 1 | 0 | 1 | + |
| F25 | 4.3205E-08 | 0 | 465 | + | 1 | 0 | 0 | = | 1 | 0 | 0 | = | 1 | 0 | 1 | + |
| F32 | 4.3205E-08 | 0 | 465 | + | 4.3205E-08 | 0 | 465 | + | 4.3205E-08 | 0 | 465 | + | 1.7344E-06 | 0 | 465 | + |
| F33 | 4.3205E-08 | 0 | 465 | + | 4.3205E-08 | 0 | 465 | + | 1 | 0 | 0 | = | 1 | 0 | 0 | = |
| F35 | 4.3205E-08 | 0 | 465 | + | 4.3205E-08 | 0 | 465 | + | 1 | 0 | 0 | = | 1 | 0 | 1 | + |
| F37 | 4.3205E-08 | 0 | 465 | + | 1 | 0 | 0 | = | 1 | 0 | 0 | = | 1 | 0 | 0 | = |
| F38 | 1.7333E-06 | 465 | 0 | - | 1.7333E-06 | 465 | 0 | - | 1.7333E-06 | 465 | 0 | - | 1.7333E-06 | 465 | 0 | - |
| F43 | 1.3670E-06 | 465 | 0 | - | 1.3670E-06 | 465 | 0 | - | 1.3670E-06 | 465 | 0 | - | 3.408E-04 | 465 | 59 | - |
| F44 | 1 | 0 | 0 | = | 1 | 0 | 0 | = | 1 | 0 | 0 | = | 1 | 0 | 0 | = |
| F45 | 1 | 0 | 0 | = | 1 | 0 | 0 | = | 1 | 0 | 0 | = | 1.7311E-06 | 0 | 465 | + |
| F47 | 1 | 0 | 0 | = | 1 | 0 | 0 | = | 1 | 0 | 0 | = | 1 | 0 | 0 | = |
| F50 | 4.3205E-08 | 0 | 465 | + | 1 | 0 | 0 | = | 1 | 0 | 0 | = | 1 | 0 | 0 | = |
| + / = / - | 13 / 7 / 7 | | | | 9 / 14 / 4 | | | | 5 / 15 / 7 | | | | 9 / 12 / 6 | | | |

**Table 4** *Continued*

| Test Functions | PSO2011 vs Modified LAB | | | | CMAES vs Modified LAB | | | | ABC vs Modified LAB | | | | JDE vs Modified LAB | | | |
|---|---|---|---|---|---|---|---|---|---|---|---|---|---|---|---|---|
| | P-value | T+ | T- | Winner | P-value | T+ | T- | Winner | P-value | T+ | T- | Winner | P-value | T+ | T- | Winner |
| F1 | 4.3205E-08 | 0 | 465 | + | 1 | 0 | 465 | + | 4.3205E-08 | 0 | 465 | + | 4.3205E-08 | 0 | 465 | + |
| F5 | 1.7311E-06 | 0 | 465 | + | 1.7311E-06 | 0 | 465 | + | 1.7331E-06 | 465 | 0 | - | 1.7311E-06 | 0 | 465 | + |
| F7 | 4.3205E-08 | 0 | 465 | + | 4.3205E-08 | 0 | 465 | + | 1 | 0 | 0 | = | 1 | 0 | 0 | = |
| F8 | 4.3205E-08 | 0 | 465 | + | 4.3205E-08 | 0 | 465 | + | 1 | 0 | 0 | = | 1 | 0 | 0 | = |
| F9 | 4.3205E-08 | 0 | 465 | + | 4.3205E-08 | 0 | 465 | + | 0.0020 | 0 | 55 | + | 4.3205E-08 | 0 | 465 | + |
| F10 | 1..920E-06 | 464 | 1 | - | 1..920E-06 | 464 | 1 | - | 0.0020 | 55 | 0 | - | 1..920E-06 | 464 | 1 | - |
| F13 | 1.6901E-06 | 465 | 0 | - | 1.6901E-06 | 465 | 0 | - | 0.0020 | 55 | 0 | - | 1.6901E-06 | 465 | 0 | - |
| F15 | 4.3205E-08 | 0 | 465 | + | 4.3205E-08 | 0 | 465 | + | 1 | 0 | 0 | = | 1 | 0 | 0 | = |
| F16 | 4.3205E-08 | 0 | 465 | + | 4.3205E-08 | 0 | 465 | + | 0.0020 | 0 | 55 | + | 4.3205E-08 | 0 | 465 | + |
| F17 | 4.3205E-08 | 0 | 465 | + | 4.3205E-08 | 0 | 465 | + | 0.0020 | 0 | 55 | + | 4.3205E-08 | 0 | 465 | + |
| F18 | 4.3205E-08 | 0 | 465 | + | 4.3205E-08 | 0 | 465 | + | 0.0020 | 0 | 55 | + | 4.3205E-08 | 0 | 465 | + |
| F19 | 0.6573 | 254 | 211 | - | 0.6573 | 254 | 211 | - | 0.0020 | 55 | 0 | - | 1.6044E-06 | 465 | 0 | - |
| F20 | 1.7217E-06 | 0 | 465 | + | 1.7217E-06 | 0 | 465 | + | 0.0020 | 0 | 55 | + | 1.7217E-06 | 0 | 465 | + |
| F21 | 1.7213E-06 | 465 | 0 | - | 1.7213E-06 | 465 | 0 | - | 0.0020 | 55 | 0 | - | 1.7213E-06 | 465 | 0 | - |
| F23 | 4.3205E-08 | 0 | 465 | + | 4.3205E-08 | 0 | 465 | + | 0.0020 | 0 | 55 | + | 4.3205E-08 | 0 | 465 | + |
| F24 | 4.3205E-08 | 0 | 465 | + | 4.3205E-08 | 0 | 465 | + | 0.0020 | 0 | 55 | + | 4.3205E-08 | 0 | 465 | + |
| F25 | 4.3205E-08 | 0 | 465 | + | 4.3205E-08 | 0 | 465 | + | 0.0020 | 0 | 55 | + | 4.3205E-08 | 0 | 465 | + |
| F32 | 4.3205E-08 | 0 | 465 | + | 4.3205E-08 | 0 | 465 | + | 0.0020 | 0 | 55 | + | 4.3205E-08 | 0 | 465 | + |
| F33 | 4.3205E-08 | 0 | 465 | + | 4.3205E-08 | 0 | 465 | + | 0.0020 | 0 | 55 | + | 4.3205E-08 | 0 | 465 | + |
| F35 | 4.3205E-08 | 0 | 465 | + | 4.3205E-08 | 0 | 465 | + | 0.0020 | 0 | 55 | + | 4.3205E-08 | 0 | 465 | + |
| F37 | 1 | 0 | 0 | = | 1 | 0 | 0 | = | 0.0020 | 0 | 55 | + | 4.3205E-08 | 0 | 465 | + |
| F38 | 1.7331E-06 | 465 | 0 | - | 1.7333E-06 | 465 | 0 | - | 0.0020 | 0 | 55 | + | 1.7333E-06 | 465 | 0 | + |
| F43 | 0.0573 | 141 | 324 | + | 0.0573 | 141 | 324 | + | 0.0020 | 0 | 55 | + | 1.3670E-06 | 465 | 0 | - |
| F44 | 1 | 0 | 0 | = | 1 | 0 | 0 | = | 0.0020 | 0 | 55 | + | 1 | 0 | 0 | = |
| F45 | 4.3205E-08 | 0 | 465 | + | 4.3205E-08 | 0 | 465 | + | 0.0020 | 0 | 55 | + | 4.3205E-08 | 0 | 465 | + |
| F47 | 1 | 0 | 0 | = | 1 | 0 | 0 | = | 0.0020 | 0 | 55 | + | 1 | 0 | 0 | = |
| F50 | 4.3205E-08 | 0 | 465 | + | 4.3205E-08 | 0 | 465 | + | 0.0020 | 0 | 55 | + | 4.3205E-08 | 0 | 465 | + |
| + / = / - | 19 / 3 / 5 | | | | 20/ 4 / 3 | | | | 19/ 5/ 3 | | | | 17/ 5/ 5 | | | |

**Table 5** Statistical pairwise comparison for CEC 2005 Benchmark test problems

| Other Algorithms vs Modified LAB | p-value | T+ | T- | Winner |
|---|---|---|---|---|
| **PSO vs Modified LAB** | 0.0442 | 34 | 119 | Modified LAB |
| **CMAES vs Modified LAB** | 0.0037 | 37 | 216 | Modified LAB |
| **ABC vs Modified LAB** | 0.2789 | 76 | 134 | Modified LAB |
| **JDE vs Modified LAB** | 0.0582 | 42 | 129 | Modified LAB |
| **CLPSO vs Modified LAB** | 0.0674 | 56 | 154 | Modified LAB |
| **SADE vs Modified LAB** | 0.3303 | 42 | 78 | Modified LAB |
| **BSA vs Modified LAB** | 0.9032 | 50 | 55 | Modified LAB |
| **IA vs Modified LAB** | 1 | 45 | 46 | Modified LAB |
| **LAB vs Modified LAB** | 1.6232E-04 | 18 | 282 | Modified LAB |

**Table 6** Statistical solutions of algorithms for CEC 2017 benchmark test problems [51 Runs] (Mean = mean solution; Std. Dev. = standard deviation; Best = best solution; Worst = worst solution)

| Functions | Statistics | LSHADE-Cn-EpsiN | FDB-SFS | LSHADE | LAB | Modified LAB |
|---|---|---|---|---|---|---|
| F1 | Mean | 8.15E+10 | 1.31E+07 | 1.14E+10 | 5.30E+10 | 5.86E+10 |
|  | Std Dev | 4.33E+08 | 1.61E+07 | 3.58E+09 | 9.61E+09 | 5.66E+09 |
|  | Best | 8.05E+10 | NA | 7.22E+09 | 2.85E+10 | 4.19E+10 |
|  | Worst | 8.21E+10 | NA | 1.88E+10 | 7.48E+10 | 7.15E+10 |
| F2 | Mean | 3.48E+04 | 2.21E+04 | 1.81E+05 | 1.17E+05 | 1.16E+05 |
|  | Std Dev | 3.75E+02 | 6.09E+03 | 4.18E+04 | 2.49E+04 | 1.47E+04 |
|  | Best | 3.37E+04 | NA | 1.06E+05 | 6.33E+04 | 7.18E+04 |
|  | Worst | 3.54E+04 | NA | 2.45E+05 | 1.83E+05 | 1.53E+05 |
| F3 | Mean | 1.21E+05 | 1.26E+02 | 1.84E+03 | 1.52E+04 | 1.67E+04 |
|  | Std Dev | 5.91E+02 | 3.62E+01 | 5.70E+02 | 5.57E+03 | 3.38E+03 |
|  | Best | 1.19E+05 | NA | 9.51E+02 | 6.97E+03 | 8.26E+03 |
|  | Worst | 1.22E+05 | NA | 3.06E+03 | 2.86E+04 | 2.34E+04 |
| F4 | Mean | 5.00E+02 | 1.08E+02 | 8.18E+02 | 7.92E+04 | 1.67E+04 |
|  | Std Dev | 5.42E-03 | 2.45E+01 | 3.19E+01 | 1.24E+04 | 3.38E+03 |
|  | Best | 5.00E+02 | NA | 7.65E+02 | 4.72E+04 | 1.67E+04 |
|  | Worst | 5.00E+02 | NA | 8.73E+02 | 1.05E+05 | 2.34E+04 |
| F5 | Mean | 6.91E+04 | 1.52E+00 | 6.58E+02 | 5.00E+02 | 5.00E+02 |
|  | Std Dev | 1.37E+04 | 8.07E-01 | 1.05E+01 | 7.85E-02 | 1.37E-02 |
|  | Best | 3.68E+04 | NA | 6.34E+02 | 5.00E+02 | 5.00E+02 |
|  | Worst | 9.95E+04 | NA | 6.78E+02 | 5.00E+02 | 5.00E+02 |
| F6 | Mean | 7.06E+02 | 1.65E+02 | 1.25E+03 | 1.72E+05 | 8.88E+04 |
|  | Std Dev | 5.51E-01 | 3.05E+01 | 7.84E+01 | 1.10E+05 | 1.70E+04 |
|  | Best | 7.04E+02 | NA | 1.09E+03 | 3.74E+04 | 4.95E+04 |
|  | Worst | 7.06E+02 | NA | 1.37E+03 | 4.08E+05 | 1.34E+05 |
| F7 | Mean | 8.74E+02 | 9.65E+01 | 1.10E+03 | 7.09E+02 | 7.05E+02 |
|  | Std Dev | 9.09E-01 | 2.10E+01 | 2.47E+01 | 7.49E+00 | 1.07E+00 |
|     | Best    | 8.71E+02  | NA       | 1.05E+03 | 7.01E+02 | 7.02E+02 |
|     | Worst   | 8.76E+02  | NA       | 1.16E+03 | 7.39E+02 | 7.08E+02 |
| F8  | Mean    | 9.34E+03  | 2.70E+02 | 8.28E+03 | 8.41E+02 | 8.43E+02 |
|     | Std Dev | 1.42E+02  | 2.54E+02 | 2.64E+03 | 1.13E+01 | 6.20E+00 |
|     | Best    | 8.94E+03  | NA       | 3.29E+03 | 8.20E+02 | 8.26E+02 |
|     | Worst   | 9.61E+03  | NA       | 1.44E+04 | 8.77E+02 | 8.59E+02 |
| F9  | Mean    | 1.95E+10  | 4.05E+03 | 9.45E+03 | 1.14E+04 | 9.21E+03 |
|     | Std Dev | 5.26E+08  | 6.12E+02 | 4.50E+02 | 1.96E+03 | 3.53E+02 |
|     | Best    | 1.75E+10  | NA       | 8.65E+03 | 6.50E+03 | 7.64E+03 |
|     | Worst   | 2.05E+10  | NA       | 1.01E+04 | 1.53E+04 | 9.82E+03 |
| F10 | Mean    | 3.17E+10  | 1.40E+02 | 9.89E+03 | 8.46E+08 | 1.18E+09 |
|     | Std Dev | 1.37E+08  | 4.41E+01 | 4.65E+03 | 8.94E+08 | 9.64E+08 |
|     | Best    | 3.15E+10  | NA       | 3.57E+03 | 5.45E+06 | 3.19E+07 |
|     | Worst   | 3.19E+10  | NA       | 1.98E+04 | 3.88E+09 | 4.25E+09 |
| F11 | Mean    | 2.78E+10  | 1.90E+06 | 6.83E+08 | 1.53E+10 | 1.64E+10 |
|     | Std Dev | 1.60E+08  | 1.96E+06 | 2.88E+08 | 3.92E+09 | 3.15E+09 |
|     | Best    | 2.73E+10  | NA       | 2.82E+08 | 6.74E+09 | 7.32E+09 |
|     | Worst   | 2.81E+10  | NA       | 1.25E+09 | 2.45E+10 | 2.23E+10 |
| F12 | Mean    | 7.37E+08  | 2.01E+04 | 1.89E+08 | 1.47E+10 | 1.55E+10 |
|     | Std Dev | 3.85E+07  | 2.57E+04 | 1.67E+08 | 5.66E+09 | 3.02E+09 |
|     | Best    | 6.48E+08  | NA       | 2.44E+07 | 5.33E+09 | 7.55E+09 |
|     | Worst   | 8.10E+08  | NA       | 6.95E+08 | 2.99E+10 | 2.14E+10 |
| F13 | Mean    | 1.71E+10  | 4.19E+03 | 1.58E+06 | 1.90E+07 | 2.25E+07 |
|     | Std Dev | 1.88E+08  | 7.55E+03 | 2.00E+06 | 1.93E+07 | 9.59E+06 |
|     | Best    | 1.63E+10  | NA       | 2.83E+04 | 1.53E+06 | 2.51E+06 |
|     | Worst   | 1.74E+10  | NA       | 8.04E+06 | 1.03E+08 | 4.77E+07 |
| F14 | Mean    | 6.46E+10  | 4.31E+03 | 1.84E+07 | 7.40E+09 | 8.25E+09 |
|     | Std Dev | 1.25E+09  | 3.54E+03 | 2.13E+07 | 2.84E+09 | 1.80E+09 |
|     | Best    | 6.07E+10  | NA       | 1.09E+06 | 3.40E+09 | 3.98E+09 |
|     | Worst   | 6.68E+10  | NA       | 9.20E+07 | 1.53E+10 | 1.25E+10 |
| F15 | Mean    | 2.45E+17  | 8.90E+02 | 4.30E+03 | 4.38E+09 | 6.51E+09 |
|     | Std Dev | 9.23E+15  | 2.72E+02 | 3.54E+02 | 4.70E+09 | 4.46E+09 |
|     | Best    | 2.09E+17  | NA       | 3.80E+03 | 1.97E+08 | 3.84E+08 |
|     | Worst   | 2.61E+17  | NA       | 5.25E+03 | 2.19E+10 | 2.40E+10 |
| F16 | Mean    | 3.74E+09  | 2.69E+02 | 2.83E+03 | 1.17E+15 | 3.26E+15 |
|     | Std Dev | 4.02E+07  | 1.63E+02 | 2.38E+02 | 1.30E+15 | 4.42E+15 |
|     | Best    | 3.66E+09  | NA       | 2.35E+03 | 8.39E+13 | 7.52E+11 |
|  | Worst | 3.82E+09 | NA | 3.14E+03 | 5.25E+15 | 2.68E+16 |
| F17 | Mean | 1.34E+17 | 9.71E+04 | 1.79E+07 | 2.93E+07 | 7.79E+07 |
|  | Std Dev | 4.77E+15 | 6.31E+04 | 3.35E+07 | 7.98E+07 | 7.82E+07 |
|  | Best | 1.21E+17 | NA | 3.38E+05 | 8.68E+04 | 1.34E+06 |
|  | Worst | 1.45E+17 | NA | 1.50E+08 | 4.91E+08 | 3.61E+08 |
| F18 | Mean | 2.60E+04 | 3.35E+03 | 3.16E+07 | 2.66E+15 | 1.43E+15 |
|  | Std Dev | 2.60E+02 | 4.82E+03 | 2.63E+07 | 8.50E+15 | 1.83E+15 |
|  | Best | 2.51E+04 | NA | 4.32E+06 | 5.12E+10 | 1.88E+11 |
|  | Worst | 2.65E+04 | NA | 1.13E+08 | 4.33E+16 | 9.95E+15 |
| F19 | Mean | 7.06E+04 | 3.36E+02 | 3.23E+03 | 1.51E+04 | 1.48E+04 |
|  | Std Dev | 2.39E+02 | 1.47E+02 | 1.92E+02 | 8.06E+03 | 3.18E+03 |
|  | Best | 6.99E+04 | NA | 2.89E+03 | 6.37E+03 | 6.81E+03 |
|  | Worst | 7.10E+04 | NA | 3.58E+03 | 3.88E+04 | 2.28E+04 |
| F20 | Mean | 9.39E+03 | 2.80E+02 | 2.60E+03 | 5.36E+04 | 5.44E+04 |
|  | Std Dev | 6.38E+01 | 4.55E+01 | 2.40E+01 | 7.12E+03 | 5.30E+03 |
|  | Best | 9.23E+03 | NA | 2.55E+03 | 3.69E+04 | 3.52E+04 |
|  | Worst | 9.50E+03 | NA | 2.63E+03 | 7.14E+04 | 6.51E+04 |
| F21 | Mean | 7.60E+04 | 1.14E+02 | 6.75E+03 | 5.71E+03 | 6.35E+03 |
|  | Std Dev | 2.31E+02 | 3.47E+00 | 2.64E+03 | 1.20E+03 | 7.40E+02 |
|  | Best | 7.54E+04 | NA | 3.14E+03 | 3.86E+03 | 4.23E+03 |
|  | Worst | 7.64E+04 | NA | 1.13E+04 | 8.49E+03 | 7.81E+03 |
| F22 | Mean | 3.79E+04 | 4.60E+02 | 2.99E+03 | 6.22E+04 | 6.44E+04 |
|  | Std Dev | 1.14E+02 | 2.44E+01 | 2.39E+01 | 7.93E+03 | 3.74E+03 |
|  | Best | 3.73E+04 | NA | 2.93E+03 | 4.14E+04 | 4.57E+04 |
|  | Worst | 3.80E+04 | NA | 3.05E+03 | 8.34E+04 | 7.13E+04 |
| F23 | Mean | 1.15E+04 | 5.20E+02 | 3.16E+03 | 3.46E+04 | 3.41E+04 |
|  | Std Dev | 1.18E+02 | 2.53E+01 | 3.85E+01 | 3.29E+03 | 1.44E+03 |
|  | Best | 1.10E+04 | NA | 3.10E+03 | 2.68E+04 | 3.02E+04 |
|  | Worst | 1.17E+04 | NA | 3.23E+03 | 4.46E+04 | 3.66E+04 |
| F24 | Mean | 4.19E+04 | 4.14E+02 | 3.48E+03 | 7.24E+03 | 7.28E+03 |
|  | Std Dev | 3.91E+02 | 1.70E+01 | 2.53E+02 | 1.25E+03 | 7.94E+02 |
|  | Best | 4.09E+04 | NA | 3.13E+03 | 5.30E+03 | 5.66E+03 |
|  | Worst | 4.26E+04 | NA | 4.04E+03 | 1.02E+04 | 8.90E+03 |
| F25 | Mean | 1.08E+04 | 1.63E+03 | 7.23E+03 | 1.89E+04 | 2.09E+04 |
|  | Std Dev | 9.19E+01 | 8.74E+02 | 4.07E+02 | 7.45E+03 | 3.82E+03 |
|  | Best | 1.05E+04 | NA | 6.64E+03 | 8.31E+03 | 1.04E+04 |
|  | Worst | 1.10E+04 | NA | 7.98E+03 | 3.49E+04 | 2.89E+04 |
| F26 | Mean | 2.55E+04 | 5.54E+02 | 3.38E+03 | 5.86E+03 | 6.27E+03 |
|  | Std Dev | 1.84E+02 | 1.98E+01 | 5.51E+01 | 5.26E+02 | 6.32E+02 |
|  | Best | 2.51E+04 | NA | 3.29E+03 | 4.55E+03 | 4.85E+03 |
|  | Worst | 2.58E+04 | NA | 3.51E+03 | 6.68E+03 | 7.64E+03 |
| F27 | Mean | 1.12E+15 | 4.78E+02 | 4.34E+03 | 9.76E+03 | 1.05E+04 |
|  | Std Dev | 5.49E+13 | 3.63E+01 | 5.09E+02 | 3.01E+03 | 2.27E+03 |
|  | Best | 9.95E+14 | NA | 3.65E+03 | 5.28E+03 | 5.57E+03 |
|  | Worst | 1.21E+15 | NA | 5.57E+03 | 1.93E+04 | 1.58E+04 |
| F28 | Mean | 4.31E+14 | 8.10E+02 | 5.11E+03 | 4.21E+13 | 4.14E+13 |
|  | Std Dev | 2.39E+13 | 1.35E+02 | 2.75E+02 | 1.02E+14 | 4.93E+13 |
|  | Best | 3.48E+14 | NA | 4.73E+03 | 3.10E+10 | 5.43E+09 |
|  | Worst | 4.62E+14 | NA | 5.81E+03 | 6.67E+14 | 2.59E+14 |
| F29 | Mean | 4.31E+14 | 6.33E+04 | 2.86E+07 | 7.23E+12 | 1.34E+13 |
|  | Std Dev | 2.12E+13 | 8.39E+04 | 1.41E+07 | 1.66E+13 | 1.93E+13 |
|  | Best | 3.78E+14 | NA | 1.07E+07 | 7.99E+09 | 2.57E+10 |
|  | Worst | 4.67E+14 | NA | 5.91E+07 | 1.00E+14 | 8.30E+13 |

**Table 7** Friedman test ranks for CEC 2017 benchmark test functions

| Friedman Test | LSHADE-Cn-EpsiN | FDB-SFS | LSHADE | LAB | Modified LAB |
|---|---|---|---|---|---|
| Mean Values | 4.1379 | 1 | 2.5517 | 3.8103 | 3.5 |
| Ranks | 5 | 1 | 2 | 4 | 3 |

## 4 Constraint Handling using Modified LAB

Most of the real-world problems are constrained in nature. These constraints can be both linear and non-linear, creating challenges for optimization algorithms. Metaheuristic algorithms are commonly used to find optimal or near-optimal solutions within constrained search spaces. However, the presence of constraints makes certain regions of the search space infeasible, limiting the algorithm's exploration. The solution is acceptable when all constraints are satisfied simultaneously. To address this, constraint handling methods are required. These techniques include penalty function methods, probability distribution-based approaches, and feasibility-based rules. The penalty function methods convert constrained problems into unconstrained ones by applying penalty values, yet selecting appropriate parameters is a problem. The niched penalty function approach (Dobnikar et al., 1999), feasibility-based rules (Ozkaya et al., 2023), and probability distribution-based techniques (Kulkarni and Shabir, 2016; Huang et al., 2023) offer alternatives. Various penalty-based methods like barrier (Luenberger and Ye, 2016; Almubarak et al., 2023), exact (Liang et al., 2023), and dynamic penalty functions (Peng et al., 2022) are efficient but suffer performance degradation as constraints increase. Augmented Lagrange Multiplier (Saeed et al., 2022), Hopfield networks (Shih and Yang, 2002), and flexible penalty functions (Curtis and Nocedal, 2008) introduce flexibility. Adaptive penalty methods (Nanakorn and Meesomklin, 2001; Han et al., 2021) adjust penalties based on fitness values and scaling. An approach for evolutionary algorithms to split penalties for individual constraints is also proposed in (Michalewicz et al., 1996) and (Coello, 2000). Other methods involving sequential penalty functions, external penalty schemes with relaxation strategies, and feasibility-based approaches that overcome the issue of local minima (Nie, 2006; Brajević, I. and Ignjatović, J., 2019). It is necessary to note that the discussed constraint handling methods are implemented in the algorithm's iterative process. In addition, these methods require information regarding the performance in the form of objective function value. Moreover, this increases computational costs when dealing with multiple or complex constraints. Therefore, it is important to explore the inherent constraint handling ability of the optimization algorithm to select the most suitable constraint handling method for it. Based on preliminary trials, it was observed that the Modified LAB is inefficient in cases where the search space was disjoint or when the optimal solution was located on the boundary of the feasible search space. To address the issue of high computational cost and to further enhance Modified LAB's unconstrained optimization ability, C-SSR method is developed and discussed in the following section. It allows the elimination of the explicit need for constraint handling between the algorithm's iterations by reducing the input search space into feasible clusters

### 4.1 Clustering-based search space reduction for constraints (C-SSR)

The objective of this method is to significantly reduce the search space while ensuring that all values which satisfy the constraints are preserved. This method is explained in the following steps:

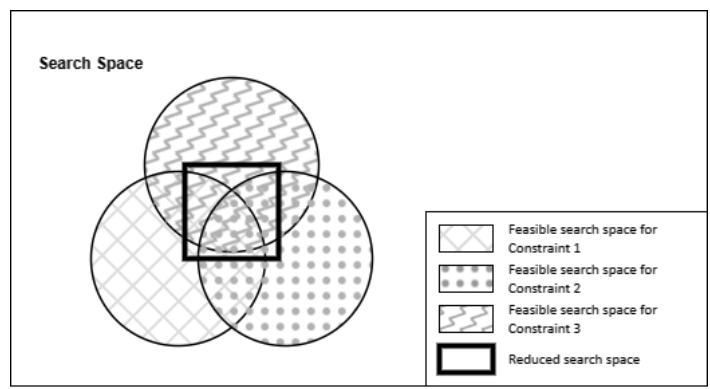

**Fig. 2** Search space overlap visualization

**Step 1: Generate points within the original search space**

The method is based on generating equidistant points for every dimension within its sampling interval. The total number of dimensions is denoted as $N$, and the number of equidistant points in each dimension as $e$ (a value intentionally minimized to reduce the computation time). The method creates a set of points, denoted as $point\_combinations$. The set of equidistant points in the $i^{th}$ dimension is defined as $P_i$ where $P_i = \{x_i^1, x_i^2, \ldots, x_i^e\}$. Consequently, the cardinality of set $P_i$ is given by $|P_i| = e$. Considering the $N$ dimensions, the method constructs the Cartesian product of these equidistant points to obtain $point\_combinations$ as follows:

$$point\_combinations = \{(x_1^p, x_2^q, \ldots, x_N^r) \mid x_1^p \in P_1, x_2^q \in P_2, \ldots, x_N^r \in P_N\}$$

where $x_i^p$ refers to the $p^{th}$ equidistant point along the $i^{th}$ dimensions. The Cartesian product is represented using the symbol | and $N$ is the total number of dimensions in the problem. Here, each element $(x_1^p, x_2^q, \ldots, x_N^r)$ in $point\_combinations$ represents a combination of values from every dimension, which effectively generates points that are spread out within the search space. This step is visualized in Fig A1 (refer section 4.2) to demonstrate the complete coverage of search space. This approach maintains reasonable computation time as the number of dimensions increases.

**Step 2: Processing of point combinations for associated constraints**

The method now applies processing of $point\_combinations$ set to obtain subsets which satisfy each constraint. This collection of subsets is denoted as $satisfying\_values$. It is represented as follows:
Let $c_i$ be the $i^{th}$ constraint, $n_i$ be the total number of points satisfying $c_i$ and let $t$ be the total number of constraints. Then, $satisfying\_values$ takes the form of a two-dimensional array:

$$satisfying\_values = \left\{ [x_1^{p_1}, x_1^{p_2}, \ldots, x_1^{p_{n_1}}], [x_2^{q_1}, x_2^{q_2}, \ldots, x_2^{q_{n_1}}], \ldots, [x_t^{r_1}, x_t^{r_2}, \ldots, x_t^{r_{n_t}}] \right\}$$

Each subset in $satisfying\_values$ corresponds to points satisfying a constraint $c_i$ ($i = 1, 2, \ldots, t$). The subset for each constraint is stored along the feasible region including its outer boundary of the constraint. Fig. 2 represents the satisfying values intersection for 3 overlapping constraints.

**Step 3: Identifying Close Points within the subsets**

K-D Tree (Bentley, 1975) is constructed for each subset in $satisfying\_values$, enabling efficient search for nearby points without traversing the entire subset. K-D Tree partitioning data points using point axes through recursive splitting, which enables efficient retrieval of points falling within a specific range or those closest to a given reference point. The maximum allowed distance ($max\_dist$) is defined to consider nearby points. The value of $max\_dist$ determines the proximity threshold for satisfying the constraint points within $satisfying\_values_i$ ($i^{th}$ subset in $satisfying\_values$ ). For each $satisfying\_values_i$, the method retrieves points from all other subsets whose distance to $satisfying\_values_i$ is within $max\_dist$ using K-D Trees. Mathematically, for any $satisfying\_values_j$ (where $j \neq i$), the method obtains the set of close points (denoted as $close\_points_{ij}$), as follows:

$$close\_points_{ij} = \{x_j^k \mid x_j^k \in satisfying\_values_j \land \exists x_i^p \in satisfying\_values_i : distance(x_i^p, x_j^k) \leq max\_dist\}$$

where $x_j^k$ represents the $k^{th}$ point in $satisfying\_values_j$, and $distance\left(x_i^p, x_j^k\right)$ calculates the distance between points $x_i^p$ and $x_j^k$. The method next combines these close points to form a comprehensive set of nearby points, referred to as $combined\_points$. The $combined\_points$ set captures regions where the optimal solution exists. Fig A3 (refer section 4.2) represents the close points for the illustrative example where only those points are retained which satisfy the $close\_points_{ij}$ criteria.

**Step 4: Creating clusters using Density-Based Spatial Clustering of Applications with Noise (DBSCAN):**

The method uses the DBSCAN (Ester et al., 1996) algorithm to discover distinct 'areas of interest' within the $combined\_points$ set. DBSCAN uses epsilon ($eps$) and the minimum number of neighbours ($MinPts$) as key parameters for the clustering process. DBSCAN starts with an arbitrary starting point and starts adding more points to a cluster which are at a maximum distance of $eps$. Those points which have at least $MinPts$ number of points around them are considered as core points of the cluster and are used to further expand the cluster, otherwise the points are treated as boundary or noise points, depending on their distance from a core point. The resulting cluster(s) represent areas where the optimal solution exists. **Fig A4** (refer section 4.2) demonstrates this step by labelling the points to the appropriate cluster according to the spatial density of the points. The use of clustering serves two main purposes:

(i) Calculate Multiple areas of Interest: Clustering reveals multiple areas, which an algorithm can explore. In real-world scenarios, various regions may satisfy all constraints, each potentially containing optimal solutions. Clustering helps to find these distinct areas for efficient exploration of search space.

(ii) Parallelization: Isolated clusters can be processed in parallel, allowing the algorithm to search for the best values independently. This parallelization strategy streamlines the search for the global optimal value and enhances the overall efficiency of the optimization process.

**Step 5: Determining new search space values from clusters:**

After obtaining the clusters of points using the DBSCAN algorithm from the $combined\_points$ set, the method proceeds to calculate the new search space values for every dimension within each cluster. For each cluster $C$, denoted as $C_1, C_2, \ldots, C_k$, where $k$ is the total number of clusters, the method iterates through the points $x_i^j$ within the cluster to determine the new input minimum and maximum values for each dimension. Let $n_c$ represent the number of points in cluster $C$, and let $i$ denote the dimension index ($i = 1, 2, \ldots, N$). Then the new search space values are calculated as follows:

$$C_{min}^i = \left(x_i^j\right) \ for \ j = 1 \ to \ n_c$$
$$C_{max}^i = \left(x_i^j\right) \ for \ j = 1 \ to \ n_c$$

The new input minimum and maximum value for dimension $i$ in Cluster $C$ ($C_{min}^i$ and $C_{max}^i \ respectively$) is obtained by finding the minimum and maximum value among all $n_c$ points within cluster $C$ along the $i^{th}$ dimension. By performing the above calculations for each cluster $C$, the method establishes the updated search space boundaries for each cluster. **Fig A5** (refer section 4.2) displays the cluster boundaries of the clusters formed in Step 4. This is further highlighted in the reduced search space cross-section present in the Fig 2, which demonstrate the possible reduction in the search space possible by the method.

## 4.2 Illustration of clustering-based search space reduction

In this section, an example is presented to demonstrate the working of the Search Space Reduction Method using Clustering using a sample 2-D constraint problem. It is important to note that although we demonstrate the steps using a 2-D problem for graphical representation, the underlying approach can be extended to a greater number of dimensions.

Sample Problem Constraints:

$C1: x^2 + y^2 \leq 1$

$C2: (x-2)^2 + y^2 \leq 1.1 \text{ or } (x+2)^2 + y^2 \leq 0.9$

Search Space Boundaries: $x \in (-5,5), y \in (-2,2)$

Working:

Step 1: Generate points within the original search space

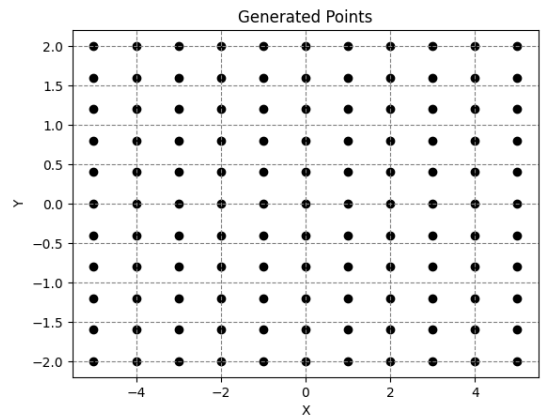

**Fig. A1** Representative points in the original search space

Step 2: Processing of $point\_combinations$ for associated constraints

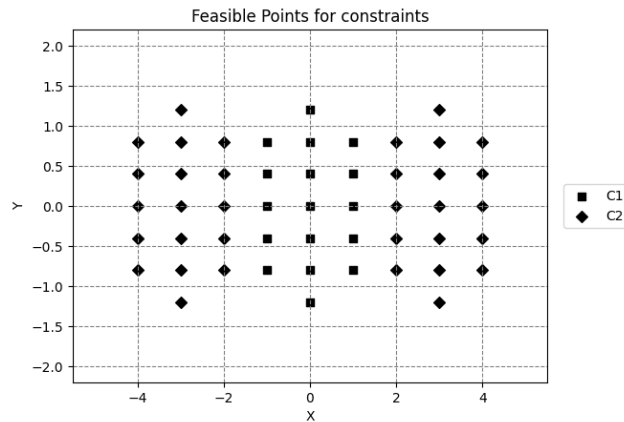

**Fig. A2** Remaining representative points after considering constraints

Step 3: Identifying Close Points within the subsets

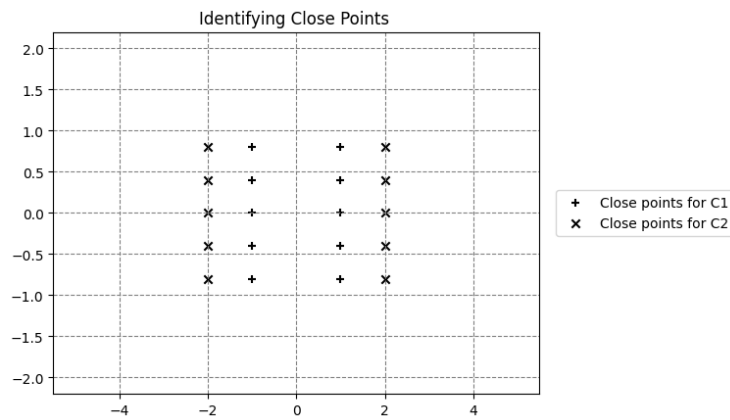

**Fig. A3** Representative points which satisfy the close point criteria

Step 4: Creating clusters using DBSCAN

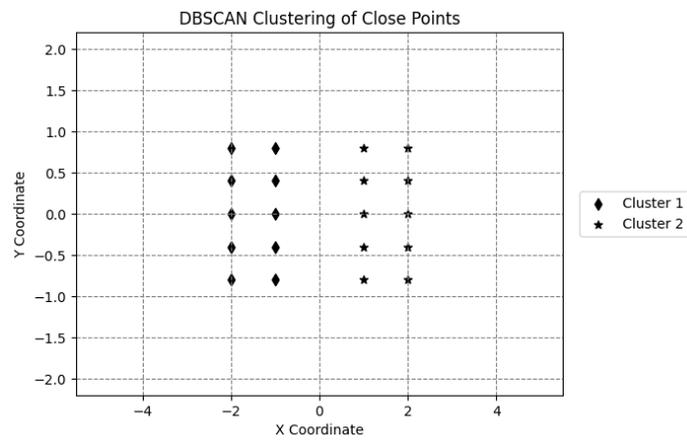

**Fig. A4** Cluster creation using spatial density of points

Step 5: Determining new search space values from Clusters

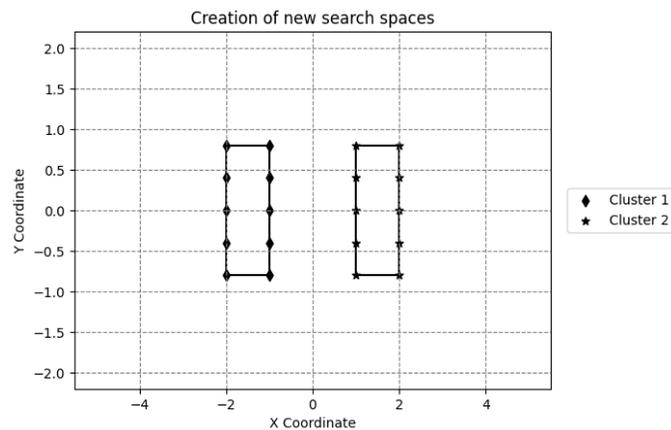

### 4.3 Constrained Modified LAB

Specific adjustments are introduced in the Modified LAB algorithm for constraints. The formulas for individual search space have been modified for leaders and advocates. Additionally, a new step has been added to check the feasibility of the generated points and to rearrange individuals iteratively. The changes are given below:

**Calculating individual search space for advocates:**

The leader to be followed by an advocate is determined using a probability-based roulette wheel approach, allowing the advocate to select a leader from their own group or other groups. The probability of a leader ($r^{L_g}$) to be selected is calculated as:

$$r^{L_g} = \frac{\frac{1}{f(x)^{L_g}}}{\sum_{g=1}^{G} \frac{1}{f(x)^{L_g}}}$$

The search space of an advocate is influenced by the leader selected through the roulette wheel approach. The search space is updated by a factor Ω. This ensures that the advocate retains its exploration capabilities while benefiting from the knowledge of a better entity:

$$\forall x_i^{A_g} \quad x_i^{A_g} = x_i^{A_g} + ((x_i^{A_g} - x_i^{r^{L_g}}) \times \Omega \times (-1)) \quad \forall A_g, i = 1,2,\ldots,N$$

where $x_i^{A_j}$ and $x_i^{r^{L_g}}$ denote the advocate and the selected leader respectively.

**Updating search space for local leaders:**

The search space of all the local leaders (except global leader) is updated by a factor Ω in the direction of the global leader as follows:

$$\forall x_i^{L_g} \quad x_i^{L_g} = x_i^{L_g} + ((x_i^{L_g} - x_i^{L^*}) \times \Omega \times (-1)) \quad \forall L_g, i = 1,2,\ldots,N$$

**Updating search space of the global leader:**

The search space of the global leader ($L^*$) is updated using the search space bounds for each dimension ($N_{min}, N_{max}$) and random value $rand \in [-1, 1]$. The step size factor ($\sigma$) is inversely proportional to the number of iterations. It is defined as follows:

$$\sigma = (\Omega/(iter + \beta/\beta))$$

where $iter$ is the current iteration count of the algorithm and β = 100. β is used here to decrease the value of $\sigma$ at slower rate allowing larger variations for global leader in earlier iterations with continued smaller changes in later iterations.

These factors are utilized in the movement of $L^*$ to introduce randomness and to ensure continued exploration in further iterations of search space. The search space is calculated as follows:

$$\forall x_i^{L^*}: \quad x_i^{L^*} = x_i^{L^*} + (abs(N_{min} - N_{max}) \times rand \times \sigma) \quad \forall L_i, i = 1,2,\ldots,N$$

**Feasibility check of the updated points:**

A feasibility check is done for the updated individuals. If, after updating the search space, any individual within the group is found to be infeasible, their variable values are not updated.

**Regrouping points after feasibility check:**

The updated individuals are ranked based on the objective function value obtained, and regrouping is performed as in **Step 1** of the Modified LAB algorithm.
Flowchart for Modified LAB algorithm for solving constrained problems is presented in Fig 3:

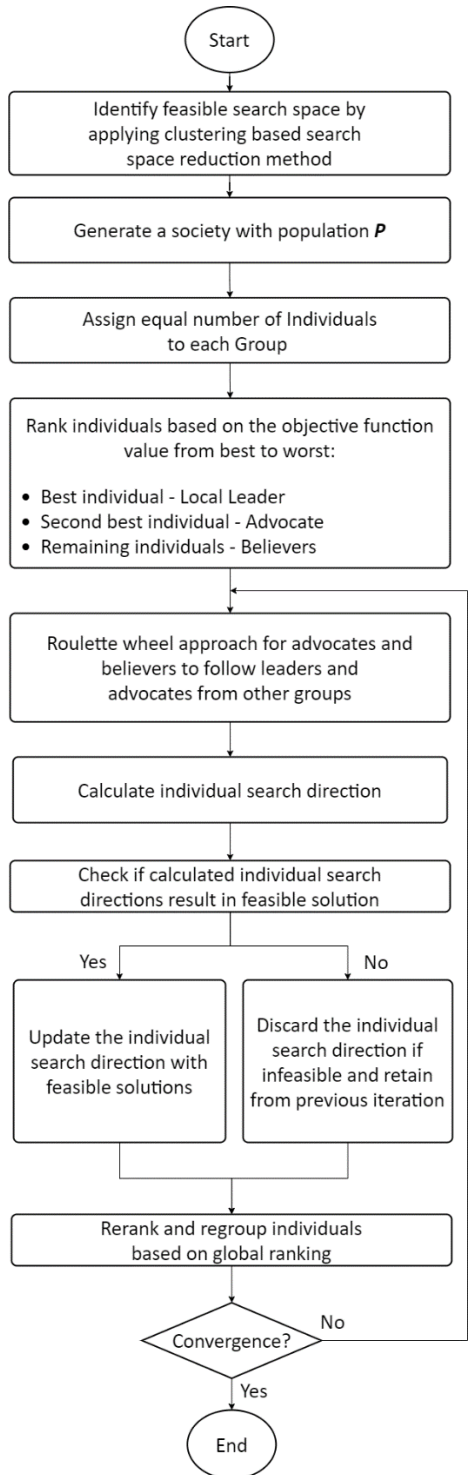

**Fig 3**: Flowchart for Modified LAB with changes for constrained problems

**5 Solutions to the Engineering Problems**

The Modified LAB's performance is evaluated on real world applications such as pressure vessel design problem, tension/compression spring design problem and welded beam design problem. The problems are referred from Minh et al., (2023). These problems consist of mixed design variables. For statistical analysis , the Modified LAB is compared against Genetic Algorithm (GA) (Coello ,2000), co-evolutionary particle swarm optimization approach (CPSO) (He and Wang, 2007), Hybrid particle swarm optimization approach (HPSO) (He and Wang, 2007) , cooperative coevolutionary method using improved augmented Lagrangian (CCiALF) (Ghasemishabankareh et al., 2016), Tunicate Swarm Algorithm (TSA) (Kaur et al., 2020), Cohort Intelligence with Panoptic Learning (CI-PL) (Krishnasamy et al., 2021) and Cohort intelligence with self-adaptive penalty function approach hybridized with colliding bodies optimization algorithm (CI-SAPF-CBO) (Kale and Kulkarni, 2021), Reptile Search Algorithm (RSA) (Abualigah et al., 2022), Golden Jackal Optimization (GJO) ( Chopra and Ansar, 2022) and Termite Life Cycle Optimizer (TLCO) (Minh et al., 2023)**.** In this evaluation, each problem is subjected to 30 separate runs for robustness assessment. The following parameters are selected for Modified LAB algorithm based on preliminary trials: individuals $n$ = 3 and groups $G$ = 71.

1. **Pressure Vessel Design Problem (Minh et al., 2023)**
   The mathematical formulation of this problem is:
   $$\text{Minimize: } f(y) = 0.6624 y_1 y_3 y_4 + 1.7781 y_2 y_3^2 + 3.1661 y_1^2 y_4 + 19.84 y_1^2 y_3$$
   Subject to:
   $$g_1(y) = -y_1 + 0.0193 y_3 \leq 0$$
   $$g_2(y) = -y_2 + 0.00954 y_3 \leq 0$$
   $$g_3(y) = -\pi y_3^2 y_4^2 - \frac{4}{3}\pi y_3^2 + 1296000 \leq 0$$
   $$g_4(y) = y_4 - 240 \leq 0$$
   $$1 \times 0.0625 \leq y_1, y_2 \leq 99 \times 0.0625, 10 \leq y_3 \leq 200, 1 \leq y_4 \leq 200$$

2. **Spring Design Problem (Minh et al., 2023)**

   The mathematical formulation of this problem is:
   $$\text{Minimize: } f(x) = (N + 2) \times d \times D$$
   Subject to:
   $$g_1(x) = 1 - \frac{N \times d^3}{71785 \times D^4} \leq 0$$
   $$g_2(x) = \frac{4 \times d^2 - D \times d}{12566 \times (d \times D^3 - D^4)} + \frac{1}{5108 \times D^2} - 1 \leq 0$$
   $$g_3(x) = 1 - \frac{140.45 \times D}{N \times d^2} \leq 0$$
   $$g_4(x) = \frac{d + D}{1.5} - 1 \leq 0$$
   $$0.05 \leq D \leq 2, 0.25 \leq d \leq 1.3, 2 \leq N \leq 15$$

3. **Welded Beam Design Problem (Minh et al., 2023)**

   The mathematical formulation of this problem is:
   $$\text{Minimize: } f(y) = 1.10471 y_2 y_1^2 + 0.04811 y_3 y_4 (14 - y_2)$$
   Subject to:
   $$g_1(y) = \tau(y) - 13000 \leq 0$$
   $$g_2(y) = \sigma(y) - 30000 \leq 0$$
   $$g_3(y) = y_1 - y_4 \leq 0$$
   $$g_4(y) = 1.10471 y_1^2 + 0.04811 y_3 y_4 (14 + y_2) - 5 \leq 0$$
   $$g_5(y) = 0.125 - y_1 \leq 0$$
   $$g_6(y) = \delta(y) - 0.25 \leq 0$$
   $$g_7(y) = 6000 - Pc(y) \leq 0$$
   $$0.1 \leq y_1 \leq 2, 0.1 \leq y_2 \leq 10, 0.1 \leq y_3 \leq 10, 0.1 \leq y_4 \leq 2$$

## 5.1 Test and Validation

Table 8 Performance comparison for Pressure Vessel Design Problem

| Design Variables | GA (Coello, 2000) | CPSO (He and Wang, 2007) | HPSO (He and Wang, 2007) | CCiALF (Behrooz et al., 2016) | TSA (Kaur et al., 2020) | CI-PL (Krishnasamy et al., 2021) | CI-SAPF-CBO (Kale and Kulkarni, 2021) | RSA (Abualigah et al., 2022) | GJO (Chopra and Ansar, 2022) | TLCO (Minh et al., 2023) | Modified LAB |
|---|---|---|---|---|---|---|---|---|---|---|---|
| $x_1$ | 8.13E-01 | 8.13E-01 | 8.13E-01 | 8.13E-01 | 7.78E-01 | 8.13E-01 | 8.13E-01 | 8.40E-01 | 7.78E-01 | 8.13E-01 | **8.13E-01** |
| $x_2$ | 4.38E-01 | 4.38E-01 | 4.38E-01 | 4.38E-01 | 3.83E-01 | 4.38E-01 | 4.38E-01 | 4.19E-01 | 3.84E-01 | 4.38E-01 | **5.63E-01** |
| $x_3$ | 4.03E+01 | 4.21E+01 | 4.21E+01 | 4.21E+01 | 4.03E+01 | 4.19E+01 | 4.21E+01 | 4.33E+01 | 4.03E+01 | 4.21E+01 | **4.08E+01** |
| $x_4$ | 2.00E+02 | 1.77E+02 | 1.77E+02 | 1.77E+02 | 2.00E+02 | 1.79E+02 | 1.77E+02 | 1.61E+02 | 2.00E+02 | 1.77E+02 | **1.99E+02** |
| $g_1(X)$ | -3.43E-02 | -1.39E-04 | -8.8×10-7 | 0.00E+00 | -9.53E-06 | -3.30E-03 | 0.00E+00 | -2.81E-03 | -8.34E-05 | 0.00E+00 | **-2.48E-02** |
| $g_2(X)$ | -5.28E-02 | -3.59E-02 | -3.59E-02 | 3.59E-02 | 3.25E+02 | -3.75E-02 | -3.59E-02 | -1.13E+03 | -1.60E+02 | -3.59E-02 | **-1.73E-01** |
| $g_3(X)$ | -2.71E+01 | -1.16E+02 | 3.12E+00 | 0.00E+00 | 1.38E-03 | 0.00E+00 | -1.5E-01 | -5.10E-03 | -1.34E-04 | 0.00E+00 | **-2.92E+04** |
| $g_4(X)$ | -4.00E+01 | -6.33E+01 | 6.34E+01 | 6.34E+01 | -4.00E+01 | -6.12E+01 | -6.34E+01 | -7.84E+01 | -4.00E+01 | -6.34E+01 | **-4.12E+01** |
| $f(X)$ | 6.29E+03 | 6.06E+03 | 6.06E+03 | 6.06E+03 | 5.87E+03 | 6.08E+03 | 6.06E+03 | 6.03E+03 | 5.89E+03 | 6.06E+03 | **6.72E+03** |

**Table 9** Statistical solutions of various algorithms for pressure vessel design problem

| Methods | Best | Mean | Worst | Std. Dev. |
|---|---|---|---|---|
| **Modified LAB** | **6.72E+03** | **7.53E+03** | **8.38E+03** | **4.48E+02** |
| TLCO (Minh et al., 2023) | 6.06E+03 | NA | NA | NA |
| RSA (Abualigah et al., 2022) | 6.03E+03 | NA | NA | NA |
| GJO (Chopra and Ansar, 2022) | 5.89E+03 | NA | NA | NA |
| CI-SAPF-CBO (Kale and Kulkarni, 2021) | 6.06E+03 | 6.06E+03 | 6.09E+03 | 9.22E+00 |
| CI-PL (Krishnasamy et al., 2021) | 6.08E+03 | 6.09E+03 | 6.09E+03 | 2.07E+00 |
| TSA (Kaur et al., 2020) | 5.87E+03 | 5.87E+03 | 5.87E+03 | 5.87E+03 |
| CCiALF (Behrooz et al., 2016) | 6.06E+03 | 6.06E+03 | 6.06E+03 | 1.01E-11 |
| HPSO (He and Wang, 2007) | 6.06E+03 | 6.10E+03 | 6.29E+03 | 8.62E+01 |
| CPSO (He and Wang, 2007) | 6.06E+03 | 6.15E+03 | 6.36E+03 | 8.65E+01 |
| GA (Coello, 2000) | 6.29E+03 | 6.29E+03 | 6.31E+03 | 7.41E+00 |

The Modified LAB algorithm is applied to solve the pressure vessel problem and the results are compared with other algorithms (refer Table 8). The algorithm reported a best, mean and worst function value of 6.72E+03, 7.53E+03 and 8.38E+03, respectively with a standard deviation 4.48E+02. It is observed that Modified LAB reported marginally worse solutions as compared to other contemporary algorithms (Table 9). This can be attributed to the large search space present in $x_3$ and $x_4$. The convergence plot for pressure vessel is presented in Fig 4.

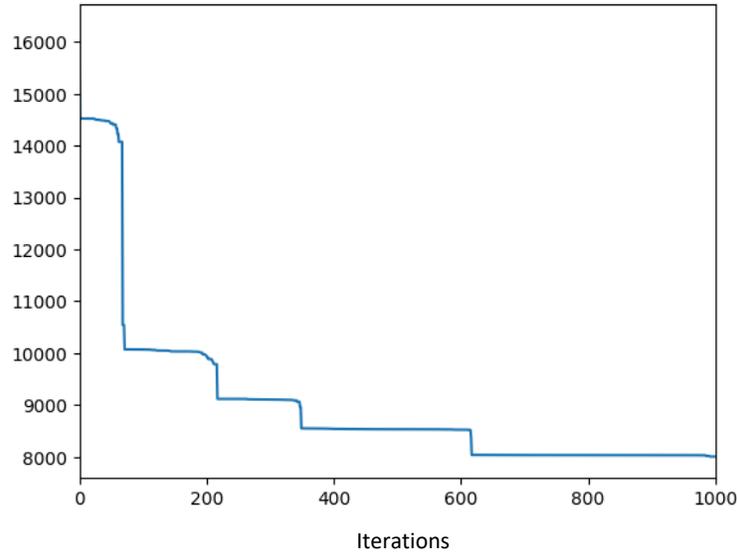

**Fig 4:** Converge plot for pressure vessel problem

Table 10 Performance comparison for Tension/Compression Spring Design Problem

| Design Variables | CAEP (Coello and Becerra, 2004) | GA (Coello, 2000) | CPSO (He and Wang, 2007) | HPSO (He and Wang, 2007) | PC (Kulkarni and Tai, 2011) | CCiALF (Behrooz et al., 2016) | TSA (Kaur et al., 2020) | CI-PL (Krishnasamy et al., 2021) | RSA (Abualigah et al., 2022) | GJO (Chopra and Ansar, 2022) | TLCO (Minh et al., 2023) | Modified LAB |
|---|---|---|---|---|---|---|---|---|---|---|---|---|
| $x_1$ | 5.00E-02 | 5.15E-02 | 5.17E-02 | 5.17E-02 | 5.06E-02 | 5.17E-02 | 5.11E-02 | 5.22E-02 | 5.80E-02 | 5.15E-02 | 5.17E-02 | **5.37E-02** |
| $x_2$ | 3.17E-01 | 3.52E-01 | 3.57E-01 | 3.58E-01 | 3.28E-01 | 3.57E-01 | 3.43E-01 | 3.70E-01 | 5.84E-01 | 3.54E-01 | 3.57E-01 | **4.04E-01** |
| $x_3$ | 1.40E+01 | 1.16E+01 | 1.13E+01 | 1.12E+01 | 1.41E+01 | 1.13E+01 | 1.21E+01 | 1.06E+01 | 4.01E+00 | 1.14E+01 | 1.13E+01 | **9.07E+00** |
| $g_1(X)$ | 0.00E+00 | -3.30E-03 | 0.00E+00 | -8.45E-04 | -5.29E-02 | 2.22E-16 | 2.72E-03 | -4.08E-10 | -1.57E-03 | -4.95E-05 | 0.00E+00 | **-5.84E-03** |
| $g_2(X)$ | -7.00E-05 | -1.00E-04 | 0.00E+00 | -1.26E-05 | -7.40E-03 | 1.11E-16 | 1.51E-03 | -4.51E-05 | 1.01E-01 | -6.33E-05 | 0.00E+00 | **-4.17E-03** |
| $g_3(X)$ | -3.97E+00 | -4.03E+00 | -4.05E+00 | -4.05E+00 | -3.70E+00 | 4.05E+00 | -4.05E+00 | -4.08E+00 | -4.91E+00 | -4.05E+00 | -4.05E+00 | **-4.09E+00** |
| $g_4(X)$ | -7.55E-01 | -7.31E-01 | -7.27E-01 | -7.27E-01 | -7.48E-01 | 7.28E-01 | -7.37E-01 | -7.19E-01 | -5.72E-01 | -7.30E-01 | -7.28E-01 | **-6.95E-01** |
| $f(X)$ | 1.27E-02 | 1.27E-02 | 1.27E-02 | 1.27E-02 | 1.35E-02 | 1.27E-02 | 1.27E-02 | 1.27E-02 | 1.18E-02 | 1.27E-02 | 1.27E-02 | **1.29E-02** |

**Table 11** Statistical solutions of various algorithms for Tension/Compression Spring Design Problem

| Methods | Best | Mean | Worst | Std. Dev. |
|---|---|---|---|---|
| **Modified LAB** | **1.29E-02** | **1.37E-02** | **1.52E-02** | **5.93E-04** |
| TLCO (Minh et al., 2023) | 1.27E-02 | NA | NA | NA |
| GJO (Chopra and Ansar, 2022) | 1.27E-02 | NA | NA | NA |
| RSA (Abualigah et al., 2022) | 1.18E-02 | NA | NA | NA |
| CI-PL (Krishnasamy et al., 2021) | 1.27E-02 | 1.28E-02 | 1.28E-02 | 2.85E-05 |
| TSA (Kaur et al., 2020) | 1.27E-02 | 1.27E-02 | 1.27E-02 | 1.01E-03 |
| CCiALF (Behrooz et al., 2016) | 1.27E-02 | 1.27E-02 | 1.27E-02 | 9.87E-08 |
| PC (Kulkarni and Tai, 2011) | 1.35E-02 | 2.61E-02 | 5.27E-02 | NA |
| HPSO (He and Wang, 2007) | 1.27E-02 | 1.27E-02 | 1.29E-02 | 5.20E-04 |
| CPSO (He and Wang, 2007) | 1.27E-02 | 1.27E-02 | 1.27E-02 | 1.58E-05 |
| GA (Coello, 2000) | 1.27E-02 | 1.28E-02 | 1.28E-02 | 3.94E-05 |
| CAEP (Coello and Becerra, 2004) | 1.27E-02 | 1.36E-02 | 1.51E-02 | 8.42E-04 |

The Modified LAB algorithm is applied to solve the tension/ compression spring design problem and the results are compared with other algorithms (refer Table 10). The algorithm reported a best, mean and worst function value 1.29E-02, 1.37E-02 and 1.52E-02, respectively with a standard deviation 5.93E-04. It is observed that Modified LAB reported comparable solutions as compared to other contemporary algorithms (Table 11). The convergence plot for tension/compression spring design problem is presented in Fig 5.

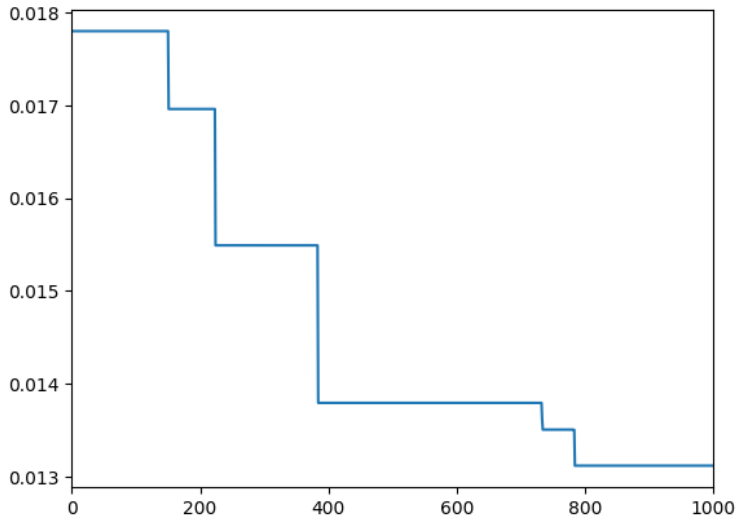

**Fig 5**: Converge plot for tension/ compression spring design problem

**Table 12** Performance comparison for Welded Beam Design Problem

| Design Variables | GA (Coello, 2000) | CAEP (Coello and Becerra, 2004) | CPSO (He et al., 2007) | HPSO (He et al., 2007) | CCiALF (Behrooz et al., 2016) | TSA (Kaur et al., 2020) | CI (Kale and Kulkarni, 2021) | CI-PL (Krishnasamy et al., 2021) | RSA (Abualigah et al., 2022) | GJO (Chopra and Ansar, 2022) | TLCO (Minh et al., 2023) | Modified LAB |
|---|---|---|---|---|---|---|---|---|---|---|---|---|
| $x_1$ | 2.09E-01 | 2.06E-01 | 2.02E-01 | 2.06E-01 | 2.06E-01 | 2.03E-01 | 2.06E-01 | 2.06E-01 | 1.44E-01 | 2.06E-01 | 2.06E-01 | **2.04E-01** |
| $x_2$ | 3.42E+00 | 3.47E+00 | 3.54E+00 | 3.47E+00 | 3.47E+00 | 3.47E+00 | 3.47E+00 | 3.47E+00 | 3.51E+00 | 3.47E+00 | 3.47E+00 | **3.60E+00** |
| $x_3$ | 9.00E+00 | 9.04E+00 | 9.05E+00 | 9.04E+00 | 9.04E+00 | 9.04E+00 | 9.04E+00 | 9.04E+00 | 8.92E+00 | 9.04E+00 | 9.04E+00 | **8.90E+00** |
| $x_4$ | 2.10E-01 | 2.06E-01 | 2.06E-01 | 2.06E-01 | 2.06E-01 | 2.01E-01 | 2.06E-01 | 2.06E-01 | 2.11E-01 | 2.06E-01 | 2.06E-01 | **2.13E-01** |
| $g_1(X)$ | -3.38E-01 | −0.00047 | −12.8397 | 1.00E-04 | 7.24E-10 | 1.66E+02 | 4.49E-02 | -7.65E-01 | 5.85E+03 | -1.32E-01 | 0.00E+00 | **-1.54E+02** |
| $g_2(X)$ | -3.54E+02 | −0.00156 | −1.24746 | -2.66E-02 | 2.00E-08 | 6.93E+02 | 9.24E-02 | -1.20E+00 | -1.02E+02 | -1.57E+01 | 0.00E+00 | **-1.61E+02** |
| $g_3(X)$ | -1.20E-03 | 0.00E+00 | −0.00149 | 0.00E+00 | 3.32E-13 | 2.14E-03 | -1.89E-12 | -5.74E-06 | -6.69E-02 | -1.00E-04 | -5.51E-14 | **-8.69E-03** |
| $g_4(X)$ | -3.41E+00 | −3.43298 | -3.42934 | -3.43E+00 | 3.43E+00 | -3.47E+00 | -3.43E+00 | -3.43E+00 | -3.41E+00 | -3.43E+00 | -3.28E+00 | **-3.39E+00** |
| $g_5(X)$ | -8.38E-02 | −0.08073 | −0.07938 | -8.07E-02 | 8.07E-02 | -7.83E-02 | -8.07E-01 | -8.07E-01 | -1.97E-02 | -8.06E-02 | -8.07E-02 | **-7.94E-02** |
| $g_6(X)$ | -2.36E-01 | −0.23554 | −0.23553 | -2.36E-01 | 2.36E-01 | -2.35E-01 | -2.36E-01 | -2.36E-01 | -2.35E-01 | -2.36E-01 | -2.36E-01 | **-2.35E-01** |
| $g_7(X)$ | -3.63E+02 | −0.00077 | −11.6813 | -2.98E-02 | 1.88E-08 | 3.92E+02 | 5.59E-02 | -1.63E-01 | -4.77E+02 | -2.81E-01 | -2.69E-03 | **-6.01E+02** |
| $f(X)$ | 1.75E+00 | 1.72E+00 | 1.73E+00 | 1.72E+00 | 1.72E+00 | 1.72E+00 | 1.72E+00 | 1.72E+00 | 1.67E+00 | 1.73E+00 | 1.72E+00 | **1.77E+00** |

**Table 13** Statistical solutions of various algorithms for Welded Beam Problem

| Methods | Best | Mean | Worst | Std. Dev. |
|---|---|---|---|---|
| **Modified LAB** | **1.77E+00** | **1.88E+00** | **2.19E+00** | **1.22E-01** |
| TLCO (Minh et al., 2023) | 1.72E+00 | NA | NA | NA |
| GJO (Chopra and Ansar, 2022) | 1.73E+00 | NA | NA | NA |
| RSA (Abualigah et al., 2022) | 1.67E+00 | NA | NA | NA |
| CI-PL (Krishnasamy et al., 2021) | 1.72E+00 | 1.72E+00 | 1.72E+00 | 8.56E-06 |
| CI (Kale and Kulkarni, 2021) | 1.72E+00 | 1.72E+00 | 1.72E+00 | 3.61E-11 |
| TSA (Kaur et al., 2020) | 1.72E+00 | 1.73E+00 | 1.73E+00 | 3.32E-03 |
| CCiALF (Behrooz et al., 2016) | 1.72E+00 | 1.72E+00 | 1.72E+00 | 5.11E-07 |
| HPSO (He et al., 2007) | 1.72E+00 | 1.75E+00 | 1.81E+00 | 4.00E-02 |
| CPSO (He et al., 2007) | 1.73E+00 | 1.75E+00 | 1.78E+00 | 1.29E-02 |
| CAEP (Coello et al., 2004) | 1.72E+00 | 1.97E+00 | 3.18E+00 | 4.43E-01 |
| GA (Coello, 2000) | 1.75E+00 | 1.77E+00 | 1.79E+00 | 1.12E-02 |

The Modified LAB algorithm is applied to solve the welded beam problem and the results are compared with other algorithms (refer Table 12). The algorithm reported a best, mean, and worst function value of 1.77E+00, 1.88E+00 and 2.19E+00 respectively with a standard deviation 1.22E-01. It is observed that Modified LAB reported marginally worse solutions as compared to other contemporary algorithms with comparable standard deviation spring design problem spring design problem (refer Table 13). The convergence plot for welded beam design problem is presented in Fig 6.

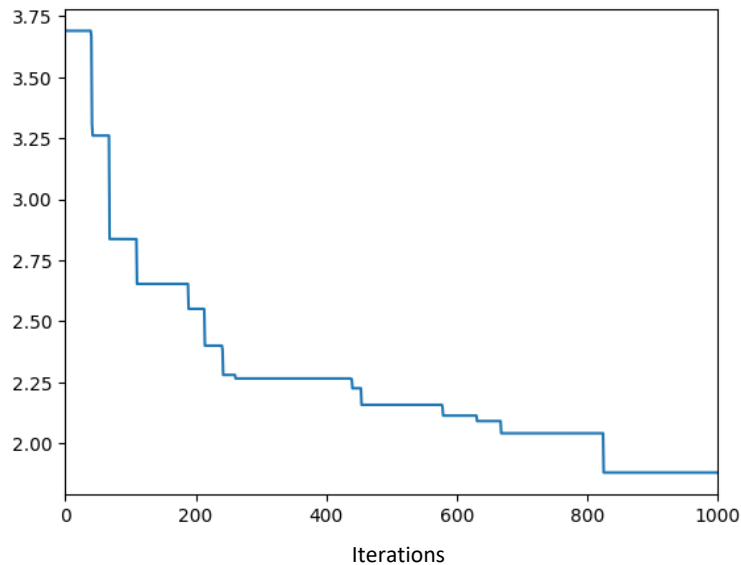

**Fig 6**: Convergence plot for welded beam problem

**6 Result Analysis and Discussion**

In the Modified LAB algorithm, two new aspects are incorporated. A sampling space reduction factor method is applied to the search space, effectively shrinking towards an optimal search space. The roulette wheel approach is also introduced, enabling believers to learn from both its own advocate and advocates from other groups. This inter-group competition allows the individuals to explore a wider search space and obtain better solutions. These improvements in the Modified LAB algorithm resulted in improved overall performance.

The Modified LAB algorithm is validated on a total of 56 benchmark test functions, 27 test problems are from CEC 2005 benchmark suite (Table 1) (Karaboga and Akay, 2009) and 29 test problems are from CEC 2017 benchmark suite (Table 2) (Wu et al., 2017). It is recommended by Awad et al. (2016) that CEC 2017 problems must be treated as black box problems. The results obtained from the Modified LAB algorithm are compared with other metaheuristic algorithms such as PSO2011 (Omran MGH, 2011), CMAES (Igel et al., 2006), ABC (Karaboga and Akay, 2009), JDE (Brest et al., 2006), CLPSO (Liang et al. 2006), SADE (Qin Yi et al., 2010), BSA (Civicioglu, 2013), IA (Huan et al., 2017), WOA (Mirjalili and Lewis, 2016), SHO (Dhiman and Kumar, 2017), AVOA (Singh et al., 2022) ,LSHADE-Cn-EpsiN (Awad et al., 2017), FDB-SFS (Aras et al., 2021), LSHADE (Mohamad et al., 2017) and LAB (Reddy et al., 2023).

The CEC 2005 benchmark test suite comprises multimodal, unimodal, separable, and non-separable functions. Two distinct statistical tests have been conducted, the Wilcoxon signed-rank test and pairwise test for the CEC 2005 benchmark test problems.

The Modified LAB algorithm demonstrated improvements in average solutions, best solutions and robustness across 66.6% (18 out of 27 test problems) of the CEC 2005 benchmark suite when compared to the LAB algorithm.
For the Wilcoxon pairwise test, a significance level of 0.05 (denoted as α) is used with the hypothesis as follows:
- Null hypothesis ($H_0$): where the median solutions of algorithms A and B are equal.
- Alternative hypothesis ($H_1$): the median solutions are not equal and that one algorithm is performed better than the other.

The last row in Table 4 provides a summary of the test results. It compares each algorithm with the Modified LAB algorithm on 27 functions of CEC 2005, using symbols +(win)/=(draw)/-(lose) to show which algorithm performed better. "+" denotes the proposed algorithm did better. "–" denotes the proposed algorithm did worse. "=" denotes there is no significant difference in performance. The Wilcoxon signed-rank test provides ranks, specifically the win (T+) and loss (T-) values, to determine if there is a meaningful difference in their performance (Civicioglu 2013) (Table 5). The Modified LAB algorithm outperformed PSO2011 (Omran MGH 2011), CMAES (Igel et al. 2006), ABC (Karaboga and Akay, 2009), JDE (Brest et al. 2006), CLPSO (Liang et al. 2006), SADE (Qin Yi et al., 2010), BSA (Civicioglu 2013) and IA (Huan et al. (2017). The Modified LAB algorithm exhibited significant enhancements in terms of both robustness and average solutions, however it also demonstrated an increase in computational time when compared to the LAB and other algorithms.

The CEC 2017 benchmark test suite consists of various types of functions, including unimodal, simple multimodal, hybrid, and composition functions. The Friedman test, a nonparametric statistical technique, is used to rank the performance of algorithms on the CEC 2017 benchmark test suite.

The solutions of CEC 2017 for Modified LAB algorithm are then compared with other recent algorithms such as LSHADE-Cn-EpSin (Awad et al., 2017), FDB-SFS (Aras et al., 2020), LSHADE (Mohamad et al., 2017) and LAB (Reddy et al., 2023). The Modified LAB algorithm performed better than LSHADE-Cn-EpsiN in terms of average and best solutions and the LAB algorithm in terms of robustness. The Modified LAB algorithm exhibited an improvement in 79.3% (23 out of the 29 test problems) of the CEC 2017 benchmark test suite compared to the LAB algorithm in terms of standard deviation. The average optimal solutions, obtained from 51 independent runs of solving the CEC 2017 benchmark test problems using the Modified LAB algorithm, are compared with the results of the CEC 2017 competition winners. The rankings in the Friedman test are based on the mean values produced by the algorithms (see Tables 7).

In numerous optimization problems, constraints are a fundamental component, requiring algorithm adaptation. To address the constraint-handling challenge, C-SSR method is devised and integrated into the Modified LAB algorithm. These adjustments yielded improvements in the following areas of constraint problem-solving:

**(a) Enhanced Search Strategy:** Instead of relying on gradient information for movement, C-SSR provides the algorithm with a narrow feasible search space. This removes the need for explicit constraint handling techniques during algorithm's iterations.

**(b) Feasibility Check:** In contrast to the original LAB algorithm, the Modified LAB algorithm incorporates a constraint feasibility check to ensure that the updated individuals lie within the feasible search space.

The Modified LAB algorithm is tested on 3 well-studied real-world engineering design problems, consisting of both continuous and discrete search spaces. The algorithm's results are compared against established benchmark algorithms, such as GA (Coello, 2000), CPSO (He and Wang, 2006) and some recent algorithms like RSA (Abualigah et al., 2022), GJO (Chopra and Ansar, 2022) and TLCO (Minh et al., 2023)**.**

The best values generated by the Modified LAB algorithm are found to be within 6% and 9% of the best values observed in the literature for the Welded Beam and Tension Compression problems respectively, with an acceptable standard deviation. This shows the algorithm's efficacy in search spaces where significant search space reduction can be achieved. However, when applied to a large search space in the Pressure Vessel problem, the algorithm's performance deteriorated, yielding best values within 14% of the observed best. It is important to note that that the observed best results for all three problems are not able to satisfy all constraints simultaneously. It is observed that the algorithm excelled at generating near-optimal variable values for discrete search space dimensions but struggled with locating optimal values for dimensions with larger input search spaces, such as $x_4$ in Pressure Vessel Problem where the search space reduction didn't substantially reduce the input space due to a considerable overlap of feasible regions.

**7 Characteristics and Limitations:**

The Modified LAB algorithm proposed here exhibited certain prominent characteristics and limitations.
These are discussed below:

- The believers in the Modified LAB algorithm can now follow advocates from other groups using the roulette wheel approach, enhancing robustness and improving exploration of the search space.

- A method for reducing the sampling space factor is introduced to update the search space of advocates and leaders, effectively narrowing it down to an optimal search space. While this approach is useful for continuous search spaces, it encounters challenges when applied to discrete or combinatorial problems.

- In cases of constrained problems, advocates also employ the roulette wheel approach to incorporate the search directions of leaders when updating their own search space. Meanwhile, local leaders utilize the search space of the global leader for updating their own search space.

- A search space reduction method using clustering (C-SSR) has been introduced as an alternative to traditional constraint handling methods for solving constrained problems. Although it allows for parallelization and faster convergence, it is less effective for equality constraints and faces scalability issues in higher dimensions.

**8 Conclusions and Future Directions**

This paper introduces an improved version of the socio-inspired LAB algorithm, referred to as the Modified LAB algorithm. It is based on the behaviour of individuals within a group, involving competition, learning, and decision-making. The Modified LAB algorithm undergoes validation by solving 56 benchmark test problems: 29 from CEC 2005 and 27 from CEC 2017. When compared to the LAB algorithm, the Modified LAB algorithm outperforms it, demonstrating improvements in average solutions and robustness in 66.6% (18 out of 27 test problems) of the CEC 2005 benchmark suite and 79.3% (23 out of 29 test problems) of the CEC 2017 benchmark suite. A statistical comparison is conducted using the two-sided Wilcoxon-signed rank test for the CEC 2005 benchmark suite problems, where the Modified LAB algorithm outperforms CLPSO, JADE, IA, PSO2011, CMAES, ABC, and JDE algorithms. While the Modified LAB algorithm shows improved robustness and average solutions, it does result in increased computational time. In the statistical Friedman test for the CEC 2017 functions, the Modified LAB algorithm outperforms both the LAB algorithm and the LSHADE-Cn-EpsiN algorithm. Additionally, the Modified LAB algorithm proves effective in solving complex problems from the CEC 2017 test suite, displaying lower standard deviations when compared to the LAB algorithm.

To solve constrained problems, the Modified LAB algorithm incorporates C-SSR, a method devised to improve its constraint-handling capabilities. This adaptation benefits the LAB algorithm in constrained problems and streamlines the computation process by reducing the search area considered by the algorithm. The approach is versatile, as the search space reduction method can be applied to various algorithms with minimal modifications. Real-world engineering problems, encompassing both continuous and discrete search spaces, are used to test the Modified LAB algorithm. Results indicate performance deviations ranging from 6% to 14% in comparison to benchmark algorithms where the optimal solutions did not satisfy all constraints simultaneously. This highlights the Modified LAB algorithm's efficacy in obtaining better solutions, as all solutions generated by it are feasible. Notably, it excels in generating near-optimal dimension values in discrete search spaces but faces challenges in larger search spaces due to limited input space reduction. Furthermore, it is essential to note that the search space reduction method, in its current implementation, struggles with equality constraints, as it requires an overlap of constraint-satisfying regions for point generation. These aspects of the algorithm's performance highlight potential areas for improvement to enhance its applicability across a broader range of constrained environments.